\documentclass[preprint,3p,times,12pt]{elsarticle}

\usepackage{amssymb}
\usepackage{amsmath}

\usepackage{algorithmic}
\usepackage{algorithm}
\usepackage{multirow}
\usepackage{url}
\usepackage{xcolor}
\usepackage{array}
\usepackage{booktabs}
\usepackage{bibunits}

\colorlet{blue}{black}
\colorlet{red}{black}

\begin{document}

\begin{frontmatter}

\title{Heterogeneous Graph Attention Network With Multimodal Feature Selection for Cancer Multiomics Integration}

\author[label1,label2]{Sina Tabakhi\corref{cor1}} %
\ead{stabakhi1@sheffield.ac.uk}
\author[label3]{Charlotte Vandermeulen} %
\ead{c.vandermeulen@sheffield.ac.uk}
\author[label3]{Ian Sudbery} %
\ead{i.sudbery@sheffield.ac.uk}
\author[label1,label2]{Haiping Lu} %
\ead{h.lu@sheffield.ac.uk}

\affiliation[label1]{organization={Centre for Machine Intelligence, University of Sheffield},%
            city={Sheffield},
            postcode={S1 4DP}, 
            country={UK}}
\affiliation[label2]{organization={School of Computer Science, University of Sheffield},%
            city={Sheffield},
            postcode={S1 4DP}, 
            country={UK}}
\affiliation[label3]{organization={School of Biosciences, University of Sheffield},%
            city={Sheffield},
            postcode={S10 2TN}, 
            country={UK}}

\cortext[cor1]{Corresponding author}

\begin{abstract}
The increase in high-dimensional multiomics data demands advanced integration models to capture the complexity of human diseases. Graph-based deep learning integration models, despite their promise, struggle with small patient cohorts and high-dimensional features, often applying independent feature selection without modeling relationships among omics. Furthermore, conventional graph-based omics models focus on homogeneous graphs, lacking multiple types of nodes and edges to capture diverse structures. We introduce a \textbf{Hetero}geneous \textbf{G}raph \textbf{AT}tention network for \textbf{omics} integration (HeteroGATomics) to improve cancer diagnosis. HeteroGATomics performs multimodal feature selection through a multi-agent system, creating dedicated networks of feature and patient similarity for each omics modality. These networks are then combined into one heterogeneous graph for learning holistic omics-specific representations and integrating predictions across modalities. {\color{red}Experiments on three cancer multiomics datasets show that HeteroGATomics outperforms baseline methods across most evaluation metrics in cancer diagnosis.} {\color{blue}External validation on an independent breast cancer cohort further demonstrates its generalizability, where HeteroGATomics outperforms graph-based multiomics integration methods across evaluation metrics.} Moreover, HeteroGATomics enhances interpretability by identifying important biomarkers contributing to the diagnosis outcomes.
\end{abstract}

\begin{keyword}
heterogeneous graph \sep graph attention networks \sep feature selection \sep multimodal learning \sep multiomics integration

\end{keyword}

\end{frontmatter}

\section{Introduction}
\label{sec:intro}
Recent advancements in sequencing technologies have significantly accelerated the generation of multimodal biological data, collectively known as multiomics. This progress enables personalized medicine by constructing comprehensive patient profiles across multiple omics modalities \cite{imani2024multi, acosta2022multimodal, zeng2025integrating}. While each omics modality contains valuable information, their collective integration enables new insights into the fundamental aspects of human disease biology, particularly in cancer research \cite{karczewski2018integrative, wang2021mogonet}. Research initiatives developing integrative multiomics models \cite{cantini2021benchmarking, acosta2022multimodal} leverage the unique and complementary characteristics of each modality to construct multimodal models that are more accurate and interpretable than unimodal models, thereby enhancing biological predictions and facilitating biomarker discovery \cite{karczewski2018integrative, li2022graph, dhillon2023biomarker}. There are three main modality fusion strategies \cite{picard2021integration, acosta2022multimodal}. \textit{Early} fusion concatenates features from different input modalities \cite{steyaert2023multimodal}, combining all modalities at the raw feature level. \textit{Joint} or \textit{intermediate} fusion processes each modality independently before integrating their low-level representations \cite{cantini2021benchmarking}, enabling more effective interactions among modalities at a lower dimensionality. \textit{Late} fusion develops separate models for each modality and merges their predictions \cite{li2024pmfn}, leveraging unimodal strengths and addressing multiomics heterogeneity. 

More recently, deep learning on graphs (or networks), referred to as graph neural networks (GNNs), has been increasingly utilized for multiomics integration \cite{li2022graph, chen2025spafusion, wang2021mogonet}. GNNs offer more accurate models with improved decision-making power by effectively capturing both the intra- and inter-omics structures within the data. Modeling each omics modality as a graph, where biological entities are nodes and their interactions are edges, facilitates a deeper understanding of entity interactions \cite{li2022graph}. Graph attention networks (GATs) are a class of GNNs that extend this capability by using an attention mechanism to learn varying importance for different interactions \cite{veličković2018graph}. This approach enhances the model's ability to focus on the most relevant biological relationships within multiomics datasets \cite{zitnik2023current}. 

GNNs in multiomics analysis, while promising, still face two key challenges. The first challenge is learning from high-dimensional data, where each omics suffers from having a large number of features compared to a small patient cohort \cite{steyaert2023multimodal, tabakhi2023multimodal, xu2025diversity}. This issue hinders the ability of GNNs to learn meaningful representations, reducing performance in biomedical applications \cite{cantini2021benchmarking}. Most current studies apply feature selection strategies independently to each omics modality \cite{picard2021integration, wang2021mogonet}, without accounting for relationships among them, which can compromise interpretability and model performance. A few studies have explored multimodal feature selection across all omics \cite{el2018min, tabakhi2022multi}, yet these often involve either a single-iteration greedy process or creating fully connected graphs rather than sparse graphs in feature space, leading to reduced predictive capability or high computational demands. Therefore, there is a clear gap in the development of better methods that effectively account for intra- and inter-omics relationships in multiomics.

The second challenge is the limited expressive power of conventional GNNs developed for homogeneous graphs. In these models \cite{zheng2024global, wang2021mogonet, wu2024mosgat}, the learning process incorporates only patient or feature similarity networks with a single type of node and edge. Representing multiple input modalities as a homogeneous graph loses crucial structural information inherent in the data, fails to account for the diverse nature of the data, and limits the model's ability to capture complex biological interactions. In contrast, heterogeneous graphs offer a solution to these limitations by modeling multiple types of nodes and edges to capture the diversity of the data. While some methods have extended learning to heterogeneous graphs \cite{zitnik2023current, ruiz2023high}, they often rely on pre-existing knowledge graphs to represent semantic relationships between different entities, such as genes, patients, and diseases. However, such pre-existing knowledge graphs may not always be available, and creating them often requires domain knowledge or expertise and can be costly \cite{chandak2023building}.

Here, we present HeteroGATomics, a novel framework employing heterogeneous GAT for integrating multiomics data for cancer diagnosis. HeteroGATomics operates in two distinct stages: multimodal feature selection and heterogeneous graph learning. Unlike previous feature selection methods, HeteroGATomics implements a multi-agent system (MAS) for a multimodal feature selection on sparse graphs, constructed across the feature spaces of all omics. This approach creates a \textit{feature similarity network} for each modality, representing one view of the input data. The proposed multimodal feature selection strategy not only achieves competitive performance but also captures structural features arising from MAS. Moreover, for each omics modality, a \textit{patient similarity network} is built, forming the second data view. We propose a dual-view approach to automatically construct modality-specific heterogeneous graphs by connecting feature and patient similarity networks, followed by representation learning with GAT encoders to generate predictions. This heterogeneous graph construction improves the expressive power of GATs by capturing diverse structural information. Leveraging late fusion for final decision-making in HeteroGATomics integrates predictions from all modalities in a supervised manner for multiomics integration \cite{wang2021mogonet}. 

We conduct comprehensive experiments to evaluate HeteroGATomics' diagnosis performance on three cancer multiomics data. {\color{red}The results show that HeteroGATomics outperforms baseline methods across most evaluation metrics}, highlighting the benefits of integrating diverse omics data. {\color{blue}To assess generalizability, we further perform an external validation in which HeteroGATomics is trained on the TCGA breast cancer cohort and tested on the independent METABRIC cohort, where it outperforms graph-based multiomics integration methods across evaluation metrics.} Additionally, we show the necessity of each module of HeteroGATomics via a series of ablation studies. We further explore HeteroGATomics' interpretability through a biomarker identification process, revealing its ability to identify cancer-related biomarkers. By pinpointing interacting biomarker networks and highlighting key cancer-related functions, it also identifies potential therapeutic targets.

\section{Methods}
\label{sec:method}
\begin{figure}[!t]
\centering
\includegraphics[width=0.9\textwidth]{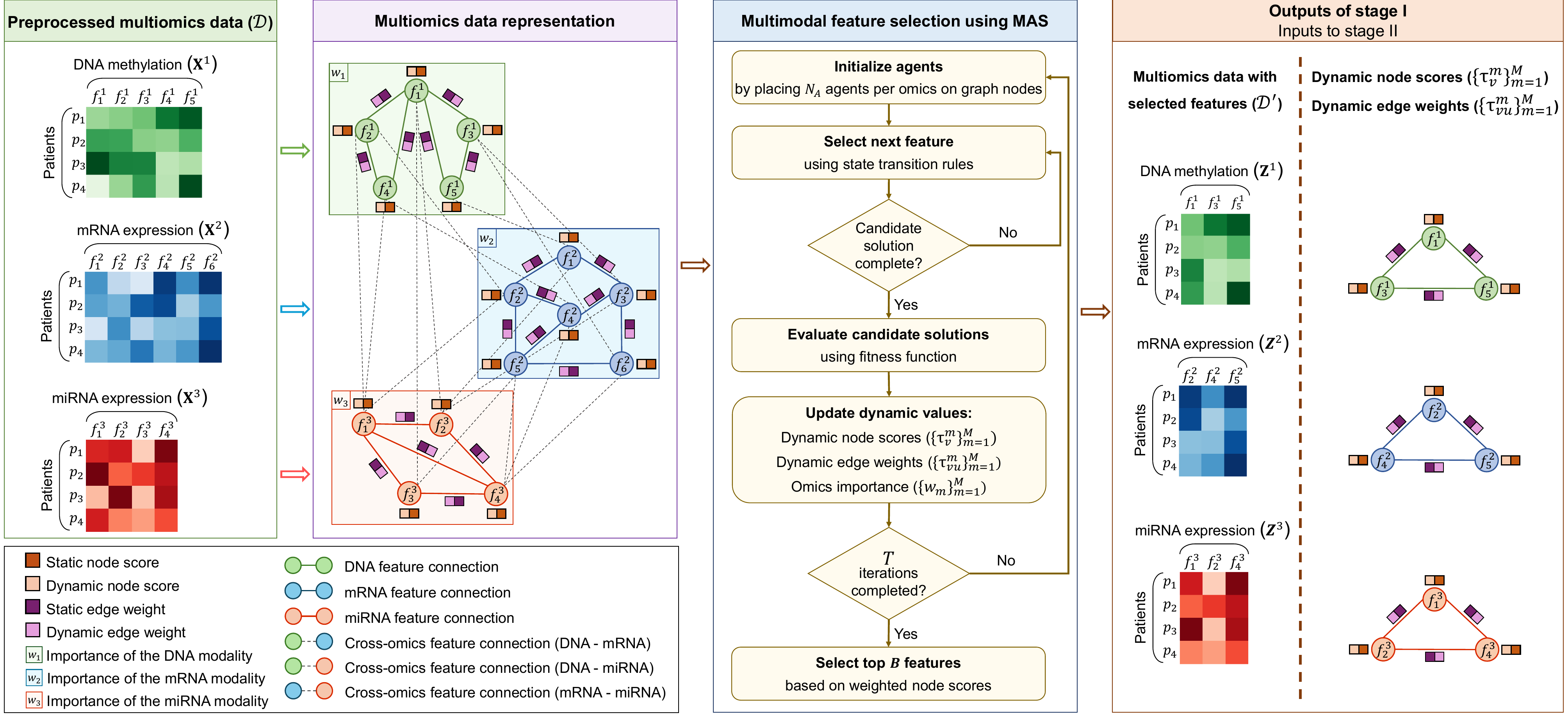}
\caption{{\color{red}
\textbf{HeteroGATomics stage I: multimodal feature selection.} 
First, HeteroGATomics represents the preprocessed multiomics data as feature similarity networks, where each network represents a specific omics modality with nodes corresponding to features and edges denoting their correlations. 
Then, a multi-agent system (MAS) selects informative features, considering both intra- and cross-modality interactions. The MAS iteratively constructs and evaluates candidate feature subsets while updating dynamic node scores, dynamic edge weights, and omics importance. After the predefined number of iterations, the MAS selects the top $B$ features based on the weighted node scores and constructs the multiomics data with the selected features for heterogeneous graph learning in stage II.}
}
\label{fig:workflow_stage_1}
\end{figure}

\begin{figure}[!t]
\centering
\includegraphics[width=0.9\textwidth]{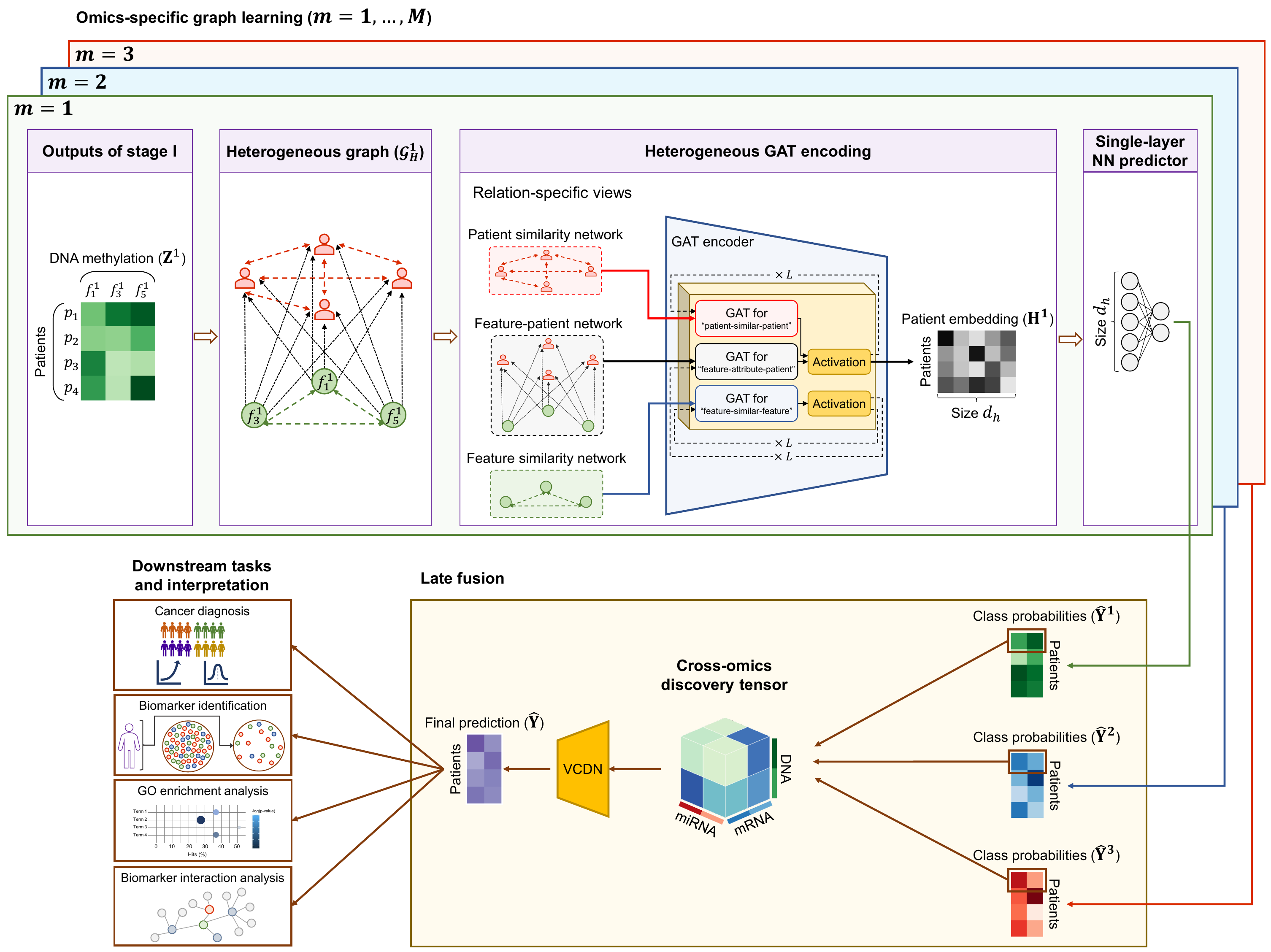}
\caption{{\color{red}
\textbf{HeteroGATomics stage II: heterogeneous graph learning}. 
First, HeteroGATomics builds a patient similarity network for each omics modality and combines it with the feature similarity network to construct a heterogeneous graph ($\mathcal{G}_H^m$). 
Then, stacked GAT layers (\textit{L}) learn heterogeneous graph representations, where each layer uses three GATs to model the three relation types in the graph. 
Afterward, a single-layer neural network predicts patient labels for each omics modality ($\hat{\mathbf{Y}}^m$). 
Finally, late fusion combines the modality-specific predictions into a cross-omics discovery tensor and feeds it into VCDN to generate the final prediction ($\hat{\mathbf{Y}}$) for downstream tasks and interpretation.}
}
\label{fig:workflow_stage_2}
\end{figure}

{\color{red}In this section, we formally introduce the overall framework of HeteroGATomics, which consists of two sequential stages that are optimized independently, as shown in Figs.~\ref{fig:workflow_stage_1} and~\ref{fig:workflow_stage_2}.
Stage I performs multimodal feature selection using a multi-agent system (MAS), while stage II performs heterogeneous graph learning and multimodal prediction using heterogeneous GAT encoders and late fusion.}

{\color{red}In stage I}, HeteroGATomics first represents each preprocessed omics as a sparse feature similarity network, where each node corresponds to an individual feature and each edge indicates the correlation between a pair of features. Then, it uses the MAS algorithm to select features in a multimodal manner from all omics modalities, leveraging both intra- and cross-modality interactions at the feature level to utilize complementary information in multiomics datasets. {\color{red}The selected features are passed to stage II for heterogeneous graph learning.}

{\color{red}In stage II}, after multimodal feature selection, HeteroGATomics creates a patient similarity network for each omics modality, where nodes represent patients and edges represent correlations between their features. Next, it constructs a heterogeneous graph for each omics modality by integrating the patient similarity network with the feature similarity network from stage I and connecting each feature node to all patient nodes. This helps capture patient-level and feature-level relationships in a unified representation, providing a comprehensive view of the dataset. Then, a heterogeneous GAT model encodes the structures inherent in each input heterogeneous graph and learns meaningful patient representations. Afterward, a single-layer fully connected neural network takes the learned patient representations as input to predict cancer. Finally, a late fusion strategy consolidates predictions across modalities by aggregating the generated predictions and feeding them into a view correlation discovery network (VCDN) \cite{wang2021mogonet} for final prediction. The model architecture and implementation details are presented in the following subsections.

To the best of our knowledge, HeteroGATomics is the first method to explore cross-modality interactions at both the feature level, using the MAS algorithm, and the label level, employing a late fusion strategy. Moreover, the multimodal feature selection algorithm in HeteroGATomics not only reduces feature dimensionality but also provides more structural information for the heterogeneous graph, benefiting downstream tasks.

\subsection{Notations}
\label{sec:not}
The important notations used in this paper are provided in Supplementary Table A.1.

\subsection{Multimodal Feature Selection} 
\label{sec:multi_fs}
The high dimensionality of multiomics datasets results in many irrelevant and redundant features, challenging GNNs to learn meaningful representations and perform classification tasks. Thus, a feature selection method is needed to identify representative features. We propose a multimodal feature selection strategy to mitigate the curse of dimensionality by modeling intra- and cross-omics interactions, instead of applying separate feature selection to each omics modality.

{\color{blue}HeteroGATomics employs a multi-agent framework for multimodal feature selection for several reasons. First, multimodal feature selection involves a large combinatorial search space, as candidate subsets must be formed from features across different omics modalities. The multi-agent design enables multiple agents to explore different candidate feature subsets rather than relying on a single search process. Second, it supports a balance between global exploration of different candidate feature subsets and local exploitation of promising features through greedy selection. Third, agents can explore features within and across omics modalities, enabling them to construct multimodal candidate solutions. Information from previously evaluated solutions is then shared among agents to guide subsequent searches, facilitating coordination across modalities. Finally, the independent construction of candidate solutions by multiple agents provides opportunities for parallel execution.}

\subsubsection{Multiomics data representation}
The input omics can be represented as interconnected graphs, with each graph corresponding to a distinct omics, as illustrated in Fig.~\ref{fig:workflow_stage_1}. 

Consider a labelled omics $\mathcal{D}^m=<\mathbf{X}^m,\mathbf{Y}>$, where $\mathbf{X}^m\in\mathbb{R}^{N\times d_m}$ represents a matrix of $N$ patients with $d_m$ features, and $\mathbf{Y}\in\mathbb{R}^{{N\times C}}$ denotes patient labels for $C$ classes, represented using one-hot encoding. Formally, each omics can be formulated as an undirected weighted graph $\mathcal{G}^m=(\mathcal{V}^m, \mathcal{E}^m, w_m)$, where $\mathcal{V}^m$ is a set of $d_m$ nodes indicating features in omics $m$, and $\mathcal{E}^m\subseteq \mathcal{V}^m\times \mathcal{V}^m$ is a set of edges connecting these features. Additionally, $\mathcal{G}^m$ includes $w_m$, denoting the relative importance of omics $m$. The graph weights are calculated as the absolute Pearson correlation between pairs of nodes, which serve as the \textit{static edge weights}, $\eta_\mathcal{E}(f_v^m, f_u^m)$. Due to the high dimensionality and the presence of correlated features within omics, utilizing a sparse graph in the feature selection algorithm accelerates training, reduces memory usage, and enhances computational efficiency. Therefore, we retain a percentage of edges with weights below the threshold $\theta_f$. Furthermore, the \textit{static node score} of each feature, $\eta_\mathcal{V}(f_v^m)$, is evaluated using ANOVA and assigned to the corresponding graph node. 

In addition to these components, we introduce a \textit{dynamic edge weight}, $\tau_{vu}^m(t)$, representing the importance of the edge between feature nodes $f_v$ and $f_u$ at time $t$, and a \textit{dynamic node score}, $\tau_v^m(t)$, indicating the significance of feature $f_v$ relative to other features in omics $m$ at time $t$. These components reflect the attractiveness of solutions and guide the algorithm toward near-optimal results. The values are initialized to small constants, $c_\mathcal{V}$ and $c_\mathcal{E}$, and assigned to each node and edge, respectively.

\subsubsection{MAS for multimodal feature selection}
Following data representation, we use MAS for multimodal feature selection, as detailed in {\color{blue}Algorithm~\ref{alg:joint_fs}}, involving an iterative improvement process with several key steps in each iteration. First, $N_A$ agents are randomly distributed across different graph nodes within each omics. Then, each agent independently constructs its solution by traversing interconnected graphs using iterative \textit{state transition rules}. These rules guide the agents to select unchosen features either probabilistically or greedily, biased by the static and dynamic node scores and edge weights. Following the construction step, solutions are evaluated using a proposed \textit{fitness function}. The highest-quality solution in the current iteration is retained as the best-found solution. 

\begin{algorithm}[!h]
\begin{footnotesize}
\caption{Multimodal Feature Selection Module}
\label{alg:joint_fs}

\color{red}
\textbf{Input} \\
\text{\hspace{0.4cm}$\mathcal{D}=<\!(\mathbf{X}^1,\mathbf{X}^2,...,\mathbf{X}^{M}),\mathbf{Y}\!>$: multiomics dataset with $M$ omics modality}\\
\text{\hspace{0.4cm}$T$: predefined number of iterations}\\
\text{\hspace{0.4cm}$N_A$: the number of agents per omics modality}\\
\textbf{Output} \\
\text{\hspace{0.4cm}$\mathcal{D^{\prime}}\!=<\!(\mathbf{Z}^1,\mathbf{Z}^2,\dots,\mathbf{Z}^{M}),\mathbf{Y}\!\!>$: reduced-dimension multiomics dataset}\\
\text{\hspace{0.4cm}${\{\tau_v^m,\tau_{vu}^m}\}_{m=1}^{M}$: dynamic node scores and edge weights}\\
\begin{algorithmic}[1]

\FOR{$m = 1$ to $M$}
    \STATE Compute static node score $\eta_\mathcal{V}(f_v^m)$
    \STATE Compute static edge weight $\eta_\mathcal{E}(f_v^m, f_u^m)$
    \STATE Initialize dynamic node score \( \tau_v^m(0) \gets c_\mathcal{V} \)
    \STATE Initialize dynamic edge weight \( \tau_{vu}^m(0) \gets c_\mathcal{E} \)
    \STATE Initialize omics importance \( w_m \gets 1/M \)
    \STATE Build a sparse graph by retaining edges with weights below \( \theta_f \)
\ENDFOR
\FOR{$t = 1$ to $T$}
    \FOR{$m = 1$ to $M$}
        \FOR{$a = 1$ to $N_A$}
            \STATE Place agent $a$ randomly on a unique node
            \STATE Build a solution by iteratively applying Eqs. (\ref{eq:trans_greedy}) and (\ref{eq:trans_prob_2}) 
            \STATE Evaluate the solution with the fitness function in Eq.~(\ref{eq:fit})
        \ENDFOR
    \ENDFOR
    \STATE Retain the solution with the highest fitness as $\mathcal{S}(\mathrm{best})$ in iteration $t$
    \STATE Update dynamic node scores using Eq.~(\ref{eq:des_upt_node})
    \STATE Update dynamic edge weights using Eq.~(\ref{eq:des_upt_edge})
    \STATE Apply omics importance updating rule
\ENDFOR
\STATE Select the top $B$ features based on weighted static and dynamic node scores
\STATE Build the multiomics dataset $\mathcal{D^{\prime}}$ with the top $B$ features
\end{algorithmic}
\end{footnotesize}
\end{algorithm}

Once all solutions are evaluated, we update the dynamic node scores and edge weights using the \textit{dynamic value updating rules}. This involves reducing the strengths of dynamic values by a constant parameter and depositing additional value on promising nodes and edges, based on their attractiveness during solution construction. Furthermore, the relative importance of each omics, denoted as $w_m$, is updated using an \textit{omics importance updating rule}, which allocates higher probability to omics whose features contributed more to the agents' solutions. These steps repeat for a predefined number of iterations. Finally, to construct reduced-dimensionality multiomics, we retain the top $B$ features selected by a weighted combination of the static and dynamic node scores. In the following sections, we describe each of these components in detail.

\paragraph{State transition rules}
Each agent constructs a solution by navigating interconnected graphs and selecting features using either a probabilistic or greedy rule. In the greedy rule, the transition from feature $f_v^m$ to feature $f_u^m$ within omics $m$ for agent $a$ is defined as:

\begin{equation}
\label{eq:trans_greedy}
f_u^m=
\underset{f_k^m \in \mathcal{J}_{v^m}^m(a)}{\arg\max}
{\left[\eta_\mathcal{V}(f_k^m)+\tau_k^m(t)+\tau_{vk}^m(t)-\textsc{Similarity}(\mathcal{S}_\mathcal{V}(a), f_k^m)\right]}, \quad \text{if } q \le q_0,
\end{equation}
where
\begin{equation}
\label{eq:trans_sim}
\textsc{Similarity}(\mathcal{S}_\mathcal{V}(a), f_k^m) =
\operatorname{Mean}_{f_v^m \in \mathcal{S}_\mathcal{V}(a)} 
\bigl[\,\eta_{\mathcal{E}}(f_v^m, f_k^m)\,\bigr],
\end{equation}
$\mathcal{J}_{v^m}^m(a)$ is the set of candidate features that agent $a$ has not yet selected within omics modality $m$, $q$ is a random number in the interval $[0, 1]$, and $q_0\in[0,1]$ is a constant parameter. In Eq.~(\ref{eq:trans_sim}), $\mathcal{S}_\mathcal{V}(a)$ denotes the set of features currently selected by agent $a$, and $\textsc{Similarity}(\mathcal{S}_\mathcal{V}(a), f_k^m)$ measures the average correlation between the candidate feature $f_k^m$ and the selected features.

In the probabilistic rule, used when $q > q_0$, an agent is allowed to probabilistically explore other omics to select any feature based on its probability value. Given that an agent $a$ is currently placed in omics $m$, the next omics $m^\prime$, which can be equal to the current omics, is randomly selected from the set of omics $\mathcal{M}$ \cite{tabakhi2022multi}. This selection is based on the set of probabilities $\mathcal{W}=\{w_1, w_2,\cdots, w_{M}\}$, indicating the relative importance of each omics. Once the next omics is chosen, the probability of transitioning to feature $f_u^{m^\prime}$ from omics $m^\prime$ is given by:

\begin{equation}\label{eq:trans_prob_2}
\operatorname*{Prob}(f_u^{m^\prime}|f_v^m)=
\frac{
\left[\eta_\mathcal{V}(f_u^{m^\prime})+\tau_u^{m^\prime}(t)-\textsc{Similarity}(\mathcal{S}_\mathcal{V}(a), f_u^{m^\prime})\right]}
{\sum\limits_{f_k^{m^\prime}\in\mathcal{J}_{v^m}^{m^\prime}(a)}\left[\eta_\mathcal{V}(f_k^{m^\prime})+\tau_k^{m^\prime}(t)-\textsc{Similarity}(\mathcal{S}_\mathcal{V}(a), f_k^{m^\prime})\right]}.
\end{equation}

In Eq.~(\ref{eq:trans_prob_2}), excluding the dynamic edge weight values simplifies the model and reduces computational cost.

\paragraph{Fitness function}
The fitness function evaluates the quality of solutions, reflecting their proximity to desired objectives. To evaluate the quality of solution $\mathcal{S}(a)$ constructed by agent $a$, we propose the following quantitative measure:

\begin{equation}
\label{eq:fit}
\textsc{Fitness}(\mathcal{S}(a))=\frac{1}{3}\Big[\textsc{Quality}(\mathcal{S}(a))+\textsc{Penalty}(\mathcal{S}(a))\Big],
\end{equation}
where
\begin{equation}
\label{eq:fit_reg}
\textsc{Penalty}(\mathcal{S}(a)) =
\operatorname{Mean}_{f_v^m \in \mathcal{S}_\mathcal{V}(a)}\!\bigl[\eta_\mathcal{V}(f_v^m)\bigr]
- \operatorname{Mean}_{(f_v^m,f_u^m) \in \mathcal{S}_\mathcal{E}(a)}\!\bigl[\eta_\mathcal{E}(f_v^m,f_u^m)\bigr],
\end{equation}
$\textsc{Quality}(\cdot)$ quantifies the classifier’s performance on the subset of selected features $\mathcal{S}(a)$. The term $\mathcal{S}_\mathcal{V}(a)$ denotes the set of selected features by agent $a$, while $\mathcal{S}_\mathcal{E}(a)$ represents the set of edges within the agent’s solution. The function $\textsc{Penalty}(\cdot)$ serves as a regularization term that penalizes irrelevant and redundant features, encouraging the discovery of more informative and representative subsets.

\paragraph{Dynamic value updating rules}
These rules aim to update the dynamic node scores and edge weights according to their contributions to high-quality solutions. These values are updated after all agents have constructed their feature subsets, as follows:

\begin{equation}
\label{eq:des_upt_node}
\tau_v^m(t+1)=\left(1-\rho_\mathcal{V}\right)\tau_v^m(t)+\rho_\mathcal{V}\left[\frac{\mathrm{Count}_\mathcal{V}(\{f_v^m\})}{\mathrm{Count}_\mathcal{V}(\mathcal{S}_\mathcal{V})}+\left[\Delta\tau_v^m(t)\right]^\mathrm{best}\right],
\end{equation}

\begin{equation} 
\label{eq:des_upt_edge}
\tau_{vu}^m(t+1)=\left(1-\rho_\mathcal{E}\right)\tau_{vu}^m(t)+
\rho_\mathcal{E}\left[\frac{\mathrm{Count}_\mathcal{E}(\{(f_{v}^m, f_{u}^m)\})}{\mathrm{Count}_\mathcal{E}(\mathcal{S}_\mathcal{E})}+\left[\Delta\tau_{vu}^m(t)\right]^\mathrm{best}\right],
\end{equation}
where parameters $\rho_\mathcal{V}, \rho_\mathcal{E}\in\left(0,1\right]$  are decay coefficients that control the balance between retaining past information and incorporating new evidence, the functions $\mathrm{Count}_\mathcal{V}(\cdot)$ and $\mathrm{Count}_\mathcal{E}(\cdot)$ determine the number of times a subset of features and edges has been selected, respectively, $\mathcal{S}_\mathcal{V}$ denotes all selected features in the current iteration, and $\mathcal{S}_\mathcal{E}$ represents all selected edges in the current iteration. In Eqs.~(\ref{eq:des_upt_node}) and~(\ref{eq:des_upt_edge}), $\bigl[\Delta\tau_v^m(t)\bigl]^{\mathrm{best}}$ and $\bigl[\Delta\tau_{vu}^m(t)\bigl]^{\mathrm{best}}$ add the fitness value of the best-found solution in the current iteration if the respective feature or edge is selected in that solution; otherwise, they are zero. 

\paragraph{Omics importance updating rule}
This rule enhances the significance of omics containing informative features for agents in subsequent iterations and is adapted from ref. \cite{tabakhi2022multi}.

\subsection{Heterogeneous Graph Learning}
\label{sec:hetero_gl}
After feature selection reduces feature dimensionality, we construct a heterogeneous graph for each omics. This graph is automatically generated, incorporating extensive structural information inherent in tabular omics for downstream tasks. Following this, a GAT encodes each heterogeneous graph, effectively representing node information. The powerful attention mechanism in the GAT architecture prioritizes important nodes and edges, which is beneficial for omics data where specific genes influence biological processes or disease mechanisms. This focus on key elements in GAT enhances performance compared to other GNN architectures \cite{veličković2018graph, hamilton2020graph}. After encoding each heterogeneous graph, a single-layer neural network performs label prediction. We combine predictions from multiple omics into a tensor representing cross-omics label correlations, which is then processed through a VCDN for final prediction. 
{\color{blue}Algorithm~\ref{alg:gat}} presents the pseudo-code for the proposed architecture.

\begin{algorithm}[!h]
\begin{footnotesize}
\caption{Heterogeneous Graph Construction and Learning Module}
\label{alg:gat}

\color{red}
\textbf{Input} \\
\text{\hspace{0.4cm}$\mathcal{D^{\prime}}=<\!\!(\mathbf{Z}^1,\mathbf{Z}^2,\dots,\mathbf{Z}^{M}),\mathbf{Y}\!\!>$: reduced-dimension multiomics dataset}\\
\text{\hspace{0.4cm}${\{\tau_v^m,\tau_{vu}^m}\}_{m=1}^{M}$: dynamic node scores and edge weights}\\
\text{\hspace{0.4cm}$T_{\text{pre}}$: number of epochs for pre-training}\\
\text{\hspace{0.4cm}$T_{\text{train}}$: number of epochs for training}\\
\textbf{Output} \\
\text{\hspace{0.4cm}${\hat{\mathbf{Y}}\in\mathbb{R}^{N\times C}}$: final predicted labels}\\
\begin{algorithmic}[1]

\FOR{$m = 1$ to $M$}
    \STATE Build patient similarity network from $\mathbf{Z}^m$
    \STATE Build feature similarity network using $\tau_v^m$, $\tau_{vu}^m$, and $\mathbf{Z}^m$
    \STATE Build heterogeneous graph $\!\mathcal{G}_H^m\!$ from feature and patient networks
\ENDFOR

\FOR{$t = 1$ to $T_{\text{pre}}$}
    \FOR{$m = 1$ to $M$}
        \STATE Apply GAT to learn representations on graph $\mathcal{G}_H^m$ via Eq.~(\ref{eq:multi-head})
        \STATE Predict omics-specific labels using a single-layer neural network
        \STATE Optimize omics-specific model parameters via Eq.~(\ref{eq:loss-func})
    \ENDFOR
\ENDFOR

\FOR{$t = 1$ to $T_{\text{train}}$}
    \FOR{$m = 1$ to $M$}
        \STATE Apply GAT to learn representations on graph $\mathcal{G}_H^m$ via Eq.~(\ref{eq:multi-head})
        \STATE Predict omics-specific labels using a single-layer neural network
    \ENDFOR
    \STATE Create a cross-omics discovery tensor for each patient via Eq.~(\ref{eq:vcdn-tensor})
    \STATE Predict final labels $\hat{\mathbf{Y}}$ with VCDN from the created tensor
    \STATE Optimize VCDN parameters by minimizing the cross-entropy loss via Eq.~(\ref{eq:vcdn-loss})
    \FOR{$m = 1$ to $M$}
        \STATE Optimize omics-specific model parameters via Eq.~(\ref{eq:loss-func})
    \ENDFOR
\ENDFOR
\end{algorithmic}
\end{footnotesize}
\end{algorithm}

\subsubsection{Heterogeneous graph construction} 
For each omics, we construct a heterogeneous graph by combining a feature similarity network with a patient similarity network (see Fig.~\ref{fig:workflow_stage_2}). The feature similarity network, deriving from the feature selection phase, consists of nodes representing selected features and edges denoting the correlations between them. Node attributes are defined by their static and dynamic node scores, along with the corresponding values from the input omics, while edge weights are determined by the static and dynamic edge weights between nodes. Complementing this, we construct the patient similarity network where nodes represent individual patients, and edges denote the dynamic edge weight between them. These correlations, quantified using the absolute Pearson correlation coefficient, serve as edge weights. Only edges with weights above a specified threshold, $\theta_s$, are maintained, which filters for the most significant patient correlations and results in a more manageable and relevant graph structure.

These two networks, indicating two distinct node types (patients and features), are interconnected to form a comprehensive heterogeneous graph encompassing three specific relations: ``\textit{patient-similar-patient},'' ``\textit{feature-similar-feature},'' and ``\textit{feature-attribute-patient}.'' Each relation in the heterogeneous graph offers unique insights: ``\textit{patient-similar-patient}'' reveals shared disease characteristics among patients; ``\textit{feature-similar-feature}'' highlights the correlated dynamics of features in biological processes; and ``\textit{feature-attribute-patient}'' provides key understanding of how individual features impact patient outcomes. This approach of relation-aware representation in the graph allows for capturing more detailed information, reflecting the diverse characteristics of the target nodes in omics datasets.

Formally, a heterogeneous graph is defined as $\mathcal{G}_H^m=(\mathcal{V}^m,\mathcal{E}^m,\mathcal{O},\mathcal{R})$, where $\mathcal{V}^m$ is the set of nodes and $\mathcal{E}^m$ is the set of edges. Each node is associated with a type via a mapping function $\phi:\mathcal{V}^m\rightarrow\mathcal{O}$, where $\mathcal{O}$ is the set of node types. Similarly, each edge is associated with a type via a mapping function $\psi:\mathcal{E}^m\rightarrow\mathcal{R}$, where $\mathcal{R}$ is the set of edge types or relations \cite{hamilton2020graph}.

\subsubsection{Representation learning and label prediction}
HeteroGATomics employs multiple GAT layers to encode omics-specific heterogeneous graphs, where each GAT within a layer is tailored to a specific relation type (as illustrated in Fig.~\ref{fig:workflow_stage_2}). For a given heterogeneous graph $\mathcal{G}_H^m$, the goal of representation learning is to develop  three relation-specific functions $f_1:\mathcal{V}_1^m\longrightarrow\mathbb{R}^{d_h}$, $f_2:\mathcal{V}_1^m\longrightarrow\mathbb{R}^{d_h}$, and $f_3:\mathcal{V}_2^m\longrightarrow\mathbb{R}^{d_h}$, each mapping the nodes involved in a particular relation into a $d_h$-dimensional embedding space. Node types are split into two disjoint sets, $\mathcal{V}^m=\mathcal{V}_1^m\cup\mathcal{V}_2^m$, where $\mathcal{V}_1^m\cap\mathcal{V}_2^m=\emptyset$. When a node is the destination of several relations, their corresponding representations are aggregated. The HeteroGATomics encoder stacks multiple layers of three individual GATs, each encoding source nodes for a specific relation. For a node $u$, layer $l+1$ aggregates messages from its relation-typed neighborhoods as follows:
\begin{equation}\label{eq:rep_lr}
\mathbf{h}_{u}^{(l+1)} =
\sigma\!\left(
\mathbf{m}_{\mathcal{N}(u)}^{(l)}
\right),
\end{equation}
where
\begin{equation}\label{eq:rep_lr_agg}
\mathbf{m}_{\mathcal{N}(u)}^{(l)} =
\frac{1}{|\mathcal{R}_u|}
\sum_{r \in \mathcal{R}_u}
\sum_{v \in \mathcal{N}_r(u)}
\alpha_{r, uv}^{(l)} \mathbf{W}_{r}^{(l)} \mathbf{h}_{v}^{(l)},
\end{equation}
$\mathcal{R}_u$ is the set of all relations for which node $u$ serves as the destination, $\mathcal{N}_r(u)$ denotes the neighbors of node $u$ under relation $r$, $\sigma(\cdot)$ represents the \textsc{LeakyReLU} nonlinearity, and $\mathbf{W}_r^{(l)}$ is the relation-specific learnable weight matrix. Self-loops are added to $\mathcal{N}_r(u)$ for both the \textit{patient–similar–patient} and \textit{feature–similar–feature} relations. In Eq.~(\ref{eq:rep_lr_agg}), $\alpha_{r, uv}^{(l)}$ denotes the normalized attention coefficient between a pair of nodes $u$ and $v$ under relation $r$, which is calculated as follows:
\begin{equation}
\label{eq:norm_att_coef}
\alpha_{r,uv}^{(l)} = \frac{\exp(\hat{\alpha}_{r,uv}^{(l)})}{\sum_{k \in \mathcal{N}_r(u)} \exp(\hat{\alpha}_{r,uk}^{(l)})},
\end{equation}
where
\begin{equation}
\label{eq:att_coef}
\hat{\alpha}_{r,uv}^{(l)} =
\sigma\!\left(
    \mathbf{a}_{r,d}^{(l)^\top} \mathbf{W}_{r,d}^{(l)}\;\mathbf{h}_u^{(l)} +
    \mathbf{a}_{r,s}^{(l)^\top} \mathbf{W}_{r,s}^{(l)}\;\mathbf{h}_v^{(l)} +
    \mathbf{a}_{r,e}^{(l)^\top} \mathbf{W}_{r,e}^{(l)}\;\mathbf{e}_{uv}
\right),
\end{equation}
$\mathbf{e}_{uv}$ indicates the edge attributes between nodes $u$ and $v$, and ${\mathbf{a}}_{r, s}^{(l)}$, ${\mathbf{a}}_{r, d}^{(l)}$, and ${\mathbf{a}}_{r, e}^{(l)}$ are the trainable attention vectors to weigh source, destination, and edge attributes, correspondingly. In Eq.~(\ref{eq:norm_att_coef}), $\exp(\cdot)$ denotes the standard exponential function and in Eq.~(\ref{eq:att_coef}), $\sigma(\cdot)$ represents the \textsc{LeakyReLU} nonlinearity.

To stabilize the attention mechanism learning process, multi-head attention \cite{veličković2018graph} is widely utilized. This scheme combines the outputs of $K$ independent attention mechanisms to form the final node representations, as follows:

\begin{equation}
\label{eq:multi-head}
\mathbf{h}_u^{(l+1)} = 
\frac{1}{K} \sum_{k=1}^K \sigma\!\big(\mathbf{m}_{\mathcal{N}(u)}^{(l,k)}\big),
\end{equation}

Once each heterogeneous graph is mapped into node embeddings through attention layers, a single-layer neural network uses them for omics-specific label prediction. We optimize trainable parameters via backpropagation to minimize the cross-entropy loss across all training patients, defined as:

\begin{equation}\label{eq:loss-func}
\mathcal{L}_m=\sum_{(\mathbf{x}_u^m, \mathbf{y}_u)\in\ \mathcal{D}_{\mathrm{tr}}^m}{{\mathrm{CE}}_m(\mathbf{y}_u,{\hat{\mathbf{y}}}_u^m)},
\end{equation}
where
$\mathcal{D}_{\mathrm{tr}}^m$ represents all the training patients in omics $m$, $\mathbf{x}_u^m$ is the $u$th training patient with its corresponding label $\mathbf{y}_u$, and ${\hat{\mathbf{y}}}_u^m$ is the class-probability vector predicted by the GAT model. The function ${\mathrm{CE}}_m(\cdot, \cdot)$ denotes the cross-entropy loss. 

We utilize VCDN, which integrates omics-specific label predictions to perform the final prediction. This network effectively learns both intra- and cross-modality correlations within the label space, thereby enhancing performance \cite{wang2021mogonet}. 
To utilize this network, we construct a cross-modality discovery tensor, $\boldsymbol{T}_u$, for each patient $u$ by integrating the predicted class probabilities from all modalities. Each element of $\boldsymbol{T}_u$ is computed as follows:

\begin{equation}\label{eq:vcdn-tensor}
\boldsymbol{T}_{u,\,t_1 t_2 \cdots t_{M}} = 
\prod_{m=1}^{M} \hat{y}_{u t_m}^{\,m}, 
\qquad t_m \in \{1, 2, \ldots, C\},
\end{equation}
where $\hat{y}_{u t_m}^{\,m}$ denotes the $t_m$th element of the predicted class-probability vector $\hat{\mathbf{y}}_u^{\,m}$ from modality $m$, $M$ is the total number of modalities, and $C$ is the number of classes. The resulting tensor $\boldsymbol{T}_u$ is reshaped into a vector of dimension $C^M$. 
VCDN, implemented as a fully connected network, takes this vector as input and outputs class probabilities for the final prediction. The network is trained using the cross-entropy loss function defined as:

\begin{equation}\label{eq:vcdn-loss}
\mathcal{L}_{\mathrm{VCDN}} =
\sum_{(\mathbf{y}_u,\, \boldsymbol{T}_u) \in \mathcal{D}_{\mathrm{tr}}}
\mathrm{CE}_{\mathrm{VCDN}}\!\left(\mathbf{y}_u,\, \mathrm{VCDN}(\boldsymbol{T}_u)\right),
\end{equation}
where $\mathrm{VCDN}(\boldsymbol{T}_u)$ is the network output for the tensor $\boldsymbol{T}_u$ of patient $u$ in the training set $\mathcal{D}_{\mathrm{tr}}$.

{\color{blue}
\subsection{Training Process}
HeteroGATomics follows two sequential stages that are optimized independently. After preprocessing the input multiomics data (Section~\ref{sec:data_process}), stage I performs multimodal feature selection using the MAS (Section~\ref{sec:multi_fs} and Algorithm~\ref{alg:joint_fs}). The MAS iteratively searches for informative feature subsets until the predefined number of iterations is reached. The multiomics data are then reduced by retaining only the selected features, and the resulting data are passed to stage II. Stage I is completed before optimization of stage II begins.

In stage II, HeteroGATomics first constructs an omics-specific heterogeneous graph by integrating the patient similarity network with the feature similarity network obtained from stage I (Section~\ref{sec:hetero_gl} and Algorithm~\ref{alg:gat}). Graph construction is completed before model training. Each omics-specific model, consisting of a GAT encoder followed by a single-layer neural network, is first pre-trained independently for $T_{\mathrm{pre}}$ epochs. The GAT encoder learns patient representations from the heterogeneous graph, while the neural network uses these representations to generate omics-specific predictions. Both components are jointly optimized by minimizing the corresponding classification loss.

After pre-training, HeteroGATomics performs multimodal training for $T_{\mathrm{train}}$ epochs. In each epoch, the omics-specific models generate patient predictions, which are combined into a cross-omics discovery tensor and passed to VCDN to generate the final prediction. The omics-specific models are optimized using their corresponding classification losses, while VCDN is optimized separately using the classification loss of the final multimodal prediction. After training, the model is evaluated using the experimental settings described in Section~\ref{sec:eval_metric}.}

\section{Experiments}
In this section, we evaluate the performance of HeteroGATomics across multiple datasets, perform ablation studies to assess the impact of individual modules, and provide a biological interpretation of the results. Implementation details are provided in Appendix B and hyperparameter tuning for HeteroGATomics is provided in Appendix C. The source code of HeteroGATomics is available in the GitHub repository.\footnote{\url{https://github.com/SinaTabakhi/HeteroGATomics}}

\subsection{Data Collection and Preprocessing}
\label{sec:data_process}
We evaluate HeteroGATomics performance across three multiomics datasets derived from The Cancer Genome Atlas (TCGA) cohort, for two downstream tasks, including bladder urothelial carcinoma (BLCA) grade classification, brain lower grade glioma (LGG) grade classification, and renal cell carcinoma (RCC) subtype classification. We download all datasets from the TCGA cohort via the UCSC Xena platform,\footnote{\url{https://xenabrowser.net/datapages/}} which provides DNA methylation (DNA), gene expression RNAseq (mRNA), and miRNA mature strand expression RNAseq (miRNA). We incorporate DNA methylation measured from the Illumina Infinium HumanMethylation450 BeadChip platform into our analysis, with 90\% of the probes from the HumanMethylation27 data present. For gene expression RNAseq, we select the IlluminaHiSeq pancan normalized version. In this modality, values were transformed using $Log2$ mean-normalized per gene across all TCGA patients. Then, only the converted data specific to the cohort of interest was extracted. Moreover, we include miRNA obtained from the IlluminaHiseq system, where values were $Log2$-transformed. More details can be found at the UCSC Xena platform.

After collecting the datasets, we perform four preprocessing steps to clean and prepare the data for machine learning models. First, only patients with available information in all omics modalities are retained. Second, features with missing values in each omics are removed. Third, all features in each omics are normalized to the range [0, 1] using min-max scaling. Fourth, features with little discriminatory power are eliminated by filtering out those with variance below a specific threshold. The threshold is set at 0.04 for DNA and mRNA, while miRNA is exempt from filtering due to its limited number of features. All datasets in the experiments use this variance threshold. Table~\ref{tbl:data} summarizes the dataset characteristics before and after the preprocessing phase. To facilitate reproducibility, the processed datasets are available in the GitHub repository.  

{\color{blue}For external validation, we additionally consider breast invasive carcinoma (BRCA) subtype classification across five classes (Normal-like, Basal-like, HER2-enriched, Luminal A, and Luminal B). The training cohort consists of 875 TCGA-BRCA patients with mRNA expression and DNA methylation modalities, obtained from the preprocessed MOGONET release~\cite{wang2021mogonet}. The independent test cohort consists of 1,272 METABRIC patients~\cite{curtis2012genomic, pereira2016somatic} with both modalities and corresponding subtype labels, downloaded from cBioPortal.\footnote{\url{https://www.cbioportal.org/}} Features are matched between the two cohorts, resulting in 972 mRNA and 458 DNA methylation features. For METABRIC, features with more than 20\% missing values are removed, and the remaining missing values are imputed using the corresponding feature mean. To account for platform differences between the cohorts, each feature is standardized to zero mean and unit variance separately within each cohort.}

\begin{table}[!t]
\centering
\scriptsize
\setlength{\tabcolsep}{2mm}
\begin{tabular}{p{0.055\linewidth} p{0.3\linewidth} p{0.085\linewidth} p{0.055\linewidth} p{0.054\linewidth} p{0.055\linewidth} p{0.045\linewidth} p{0.045\linewidth} p{0.06\linewidth}} 
\hline
\multirow{2}{*}{Dataset} & \multirow{2}{*}{Categories} & \multirow{2}{=}{\#Patients} & \multicolumn{3}{p{0.185\linewidth}}{\#Original features} & \multicolumn{3}{p{0.2\linewidth}}{\#Preprocessed features} \\ 
\cline{4-9} 
& & & DNA & mRNA & miRNA & DNA & mRNA & miRNA \\ 
\hline
BLCA & 
High-grade: 397, Low-grade: 21 & 418 & 485,577 & 20,530 & 2,210 & 7,999 & 2,373 & 249 \\
LGG & Grade II: 254, Grade III: 268 & 522 & 485,577 & 20,530 & 2,157 & 8,277 & 1,166 & 287 \\
RCC & KICH: 65, KIRC: 201, KIRP: 294 & 560 & 485,577 & 20,530 & 1,847 & 4,107 & 2,456 & 238 \\
\hline
\end{tabular}
\caption{A Summary of multiomics data characteristics.\label{tbl:data}}
\end{table}

\subsection{Baselines}
We compare the performance of the feature selection module of HeteroGATomics (that is, HeteroGATomics\textsubscript{MAS}) with five baseline methods: (1) mutual information (MI) quantifies the dependency between two random variables, yielding a non-negative value that is often used to select features with the highest information shared with the target class; (2) recursive feature elimination (RFE) \cite{guyon2002gene} recursively removes the least important features based on an estimator's weights, starting with all features and stopping at the desired number; (3) minimal-redundancy–maximal-relevance (mRMR) \cite{mrmr2005peng} selects features that have the highest relevance with the target class and are minimally redundant with each other, balancing the trade-off between relevance and redundancy; (4) minimal-redundancy–maximal-relevance multi-view (mRMR-mv) \cite{el2018min} is an adaption of mRMR to multiomics integration setting; (5) multi-agent architecture for multi-omics (MAgentOmics) \cite{tabakhi2022multi} is an iterative improvement method that extends the MAS algorithm to work with multiomics. Notably, mRMR-mv and MAgentOmics are specifically designed for multimodal feature selection within multiomics datasets. To evaluate the performance of MI, RFE, and mRMR, we concatenate the selected features from each modality to serve as the input for a classifier.

Moreover, we compare the classification performance of HeteroGATomics with that of {\color{red}ten} multiomics integration methods. We adopt early fusion using five classifiers -- $k$-nearest neighbors (KNN), multilayer perceptron (MLP), random forest (RF), Ridge regression (Ridge), gradient tree boosting (XGBoost) -- concatenating 100 features from each modality selected by MI. We also utilize multimodal feature selection techniques by mRMR-mv and MAgentOmics, specifically tailored for multiomics data integration, each selecting 300 features collectively from all modalities based on their respective strategies. {\color{blue}Moreover, we employ the recently developed PEARL \cite{zhao2026pearl}, a graph-based multi-omics integration method for classification, which constructs omics-specific patient similarity networks using weighted Pearson correlation, employs spectral graph convolutional networks for omics-specific feature learning, and integrates the learned representations using pooling or concatenation followed by an MLP.} In addition, we employ the multi-omics method MOSGAT \cite{wu2024mosgat} for cancer classification, which leverages specificity-aware GATs for omics-specific learning and a cross-modal attention mechanism to capture inter-omics associations. We also use multi-omics graph convolutional networks (MOGONET) \cite{wang2021mogonet}, a supervised late fusion method for classification tasks, which leverages graph convolutional networks (GCN) for omics-specific patient classification and employs VCDN to combine initial predictions from each omics modality into a final label. For all three methods, 100 features from each modality are selected using MI for a fair comparison.

\subsection{Evaluation Strategies and Metrics}
\label{sec:eval_metric}
We evaluate the classification performance on three datasets using stratified 10-fold cross-validation. This approach ensures robust performance evaluation by splitting each dataset into training and test sets while maintaining a balanced class representation. In each fold, nine sets are used for training, further split into training and validation sets at a 9:1 ratio, while the remaining set is used for testing. This process is repeated ten times, ensuring each set is used for testing exactly once. {\color{red}For each fold, feature selection is performed using only the training data and labels, and the selected features are then used for heterogeneous graph learning within the same fold. The same training and test indices are maintained across both stages of HeteroGATomics.} We report the mean and the standard deviation of the evaluation metrics calculated on the test sets across experiments. For the hyperparameter configuration of HeteroGATomics, we report the mean of the evaluation metrics calculated on the validation sets. To ensure a fair comparison across different methods, we maintain identical splits for all evaluations. {\color{blue}For external validation, all methods are trained on the full TCGA-BRCA cohort and evaluated once on the METABRIC cohort. The METABRIC cohort is not used for feature selection or model selection. Baseline methods are run using their published BRCA settings.}

To evaluate model performance for binary classification tasks, we use {\color{red}eight} metrics, including area under the receiver operating characteristic curve (AUROC), accuracy, {\color{blue}balanced accuracy (BACC), Matthews correlation coefficient (MCC)}, negative predictive values (NPVs), positive predictive values (PPVs), sensitivity, and specificity. For multi-class classification, we use six metrics, including accuracy, macro-averaged F1 score (Macro F1), micro-averaged F1 score (Micro F1), weighted-averaged F1 score (Weighted F1), precision, and recall. 

\subsection{Classification Performance Comparison {\color{blue}Within Cohorts}}
We compare the classification performance of HeteroGATomics with multiomics integration methods. Tables~\ref{tbl:class_all_blca}--\ref{tbl:class_all_rcc} present the detailed classification comparisons for the BLCA, LGG, and RCC datasets. From Tables~\ref{tbl:class_all_blca} and~\ref{tbl:class_all_lgg}, we observe that HeteroGATomics outperforms baseline multiomics integration methods for the binary classification task on BLCA and LGG in terms of AUROC, accuracy, \textcolor{blue}{BACC, MCC,} NPV, PPV, sensitivity, and specificity, except for \textcolor{blue}{sensitivity and} specificity in BLCA, and NPV and sensitivity in LGG. HeteroGATomics still achieves the second-best \textcolor{blue}{in sensitivity for BLCA and} in NPV for LGG. Furthermore, Table~\ref{tbl:class_all_rcc} demonstrates HeteroGATomics' superiority for the multi-class classification task on RCC, excelling in all the evaluated metrics. {\color{blue}Wilcoxon signed-rank tests \cite{demvsar2006statistical} across folds further show statistically significant improvements of HeteroGATomics in several comparisons, particularly on BLCA and LGG (see Appendix~E).}

The outstanding performance of HeteroGATomics indicates that the combined power of its HeteroGATomics\textsubscript{MAS} module and HeteroGATomics\textsubscript{GAT} module trained on generated heterogeneous graphs significantly enhances the capabilities of a deep learning model for multiomics integration. 

\begin{figure}[!t]
\color{blue}
\centering
    \includegraphics[width=1.0\textwidth]{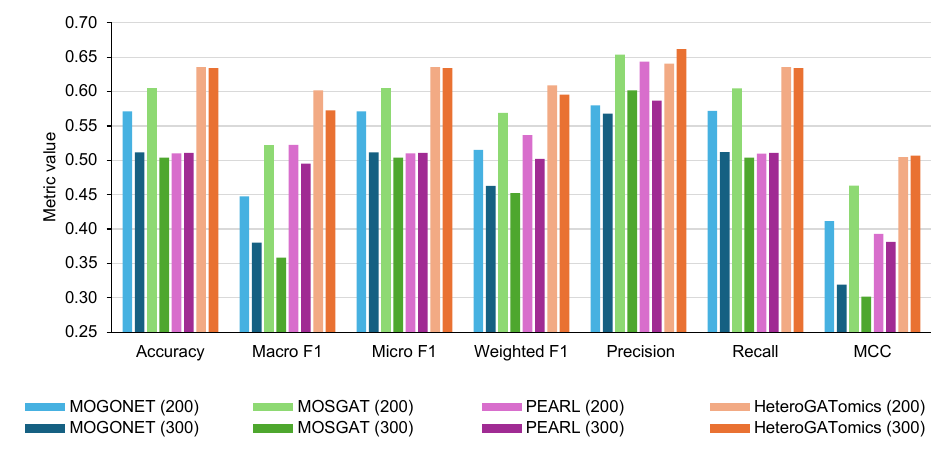}
    \caption{External validation on METABRIC. Models are trained on TCGA-BRCA (n = 875) and tested on METABRIC (n = 1,272) for five-class breast cancer subtype classification using mRNA expression and DNA methylation. Numbers in parentheses indicate the total number of selected features across both modalities.}
    \label{fig:external_val}
\end{figure}

{\color{blue}
\subsection{External Validation on an Independent Cohort}
To assess generalizability beyond the three cancer types evaluated in the within-cohort experiments, we additionally use breast cancer for external validation, as TCGA-BRCA and the independent METABRIC cohort share both omics modalities and the same five subtype classes. All methods are trained on TCGA-BRCA and evaluated on METABRIC (Fig.~\ref{fig:external_val}). We focus the comparison on graph-based multiomics integration methods (MOGONET, MOSGAT, and PEARL), as these methods, like HeteroGATomics, exploit patient relationships and are therefore directly affected by cross-cohort differences in patient similarity structure.

With 200 selected features, HeteroGATomics improves over the best-performing baseline for each metric by 5.06\% in accuracy, 15.2\% in Macro F1, 5.06\% in Micro F1, 7.03\% in Weighted F1, 5.13\% in recall, and 8.92\% in MCC. Precision is the only metric for which MOSGAT and PEARL achieve marginally higher values at this feature size. HeteroGATomics also shows greater robustness to the number of selected features, with only minor performance changes between 200 and 300 features, whereas the baseline methods generally decline. These results demonstrate that HeteroGATomics maintains strong cross-cohort performance despite differences in measurement platforms between TCGA-BRCA and METABRIC.
}

\begin{table}[!t]
\centering
\scriptsize
\setlength{\tabcolsep}{1.1mm}
\begin{tabular}{lll>{\color{blue}}l>{\color{blue}}lllll}
\hline
Method         & AUROC               & Accuracy              & BACC              & MCC          & NPV                 & PPV                 & Sensitivity         & Specificity         \\ 
\hline
KNN            & 0.779 $\pm$ 0.191   & 0.955 $\pm$ 0.027    & 0.779 $\pm$ 0.191 & 0.517 $\pm$ 0.331 & 0.978 $\pm$ 0.020 & 0.517 $\pm$ 0.329    & 0.583 $\pm$ 0.382 & 0.975 $\pm$ 0.020 \\
MLP            & 0.744 $\pm$ 0.194   & \underline{0.962 $\pm$ 0.026} &    0.744 $\pm$ 0.194 & 0.513 $\pm$ 0.368 & 0.973 $\pm$ 0.023    & 0.583 $\pm$ 0.417    & 0.500 $\pm$ 0.387    & 0.987 $\pm$ 0.017 \\
RF             & 0.684 $\pm$ 0.149   & 0.955 $\pm$ 0.022    & 0.684 $\pm$ 0.149 & 0.444 $\pm$ 0.317 & 0.968 $\pm$ 0.016    & 0.600 $\pm$ 0.436    & 0.383 $\pm$ 0.299    & 0.985 $\pm$ 0.020 \\
Ridge          & 0.720 $\pm$ 0.129   & \underline{0.962 $\pm$ 0.016} &  0.720 $\pm$ 0.129   & 0.513 $\pm$ 0.271 & 0.971 $\pm$ 0.018    & \underline{0.650 $\pm$ 0.391} & 0.450 $\pm$ 0.269    & \underline{0.990 $\pm$ 0.017} \\
XGBoost        & 0.760 $\pm$ 0.175   & \textbf{0.964 $\pm$ 0.016} &  0.760 $\pm$ 0.175  & \underline{0.552 $\pm$ 0.302} & 0.976 $\pm$ 0.019    & \underline{0.650 $\pm$ 0.369} & 0.533 $\pm$ 0.356    & 0.987 $\pm$ 0.013 \\
mRMR-mv        & 0.683 $\pm$ 0.146   & 0.952 $\pm$ 0.018    & 0.683 $\pm$ 0.146 & 0.402 $\pm$ 0.288 &  0.968 $\pm$ 0.015    & 0.500 $\pm$ 0.387    & 0.383 $\pm$ 0.299    & 0.982 $\pm$ 0.020 \\
MAgentOmics    & 0.572 $\pm$ 0.114   & 0.952 $\pm$ 0.015    & 0.572 $\pm$ 0.114 & 0.184 $\pm$ 0.294 &  0.957 $\pm$ 0.014    & 0.250 $\pm$ 0.403    & 0.150 $\pm$ 0.229    & \textbf{0.995 $\pm$ 0.010} \\
MOGONET        & 0.884 $\pm$ 0.160   & 0.948 $\pm$ 0.023    & 0.617 $\pm$ 0.164 & 0.244 $\pm$ 0.336 &  0.961 $\pm$ 0.019    & 0.292 $\pm$ 0.407    & 0.250 $\pm$ 0.335    & 0.985 $\pm$ 0.023 \\
MOSGAT         & \textbf{0.961 $\pm$ 0.029}	  & 0.955 $\pm$ 0.017	 &  0.669 $\pm$ 0.155  & 0.352 $\pm$ 0.312 & 0.966 $\pm$ 0.019	& 0.400 $\pm$ 0.382	& 0.350 $\pm$ 0.320	& 0.987 $\pm$ 0.017 \\
\color{blue}PEARL          & \color{blue}\underline{0.959 $\pm$ 0.060}	  & \color{blue}0.818 $\pm$ 0.057	 & \color{blue}\underline{0.809 $\pm$ 0.124} & \color{blue}0.368 $\pm$ 0.111 & \color{blue}\underline{0.982 $\pm$ 0.015}	& \color{blue}0.244 $\pm$ 0.086	& \color{blue}\textbf{0.800 $\pm$ 0.245}	& \color{blue}0.819  $\pm$ 0.060 \\
HeteroGATomics & \textbf{0.961 $\pm$ 0.065} & \textbf{0.964 $\pm$ 0.027} & \textbf{0.823 $\pm$ 0.156} & \textbf{0.636 $\pm$ 0.296} & \textbf{0.983 $\pm$ 0.016} & \textbf{0.673 $\pm$ 0.360} & \underline{0.667 $\pm$ 0.316} & 0.980 $\pm$ 0.027 \\
\hline
\end{tabular}
\caption{Classification performance comparison (mean $\pm$ standard deviation over 10-fold cross-validation) for the BLCA dataset. The best performance for each metric is in bold; second-best results are underlined.}
\label{tbl:class_all_blca}
\end{table}

\begin{table}[!t]
\centering
\scriptsize
\setlength{\tabcolsep}{1.1mm}
\begin{tabular}{lll>{\color{blue}}l>{\color{blue}}lllll}
\hline
Method         & AUROC                & Accuracy             & BACC              & MCC                  & NPV                  & PPV                  & Sensitivity         & Specificity         \\ \hline
KNN            & 0.670 $\pm$ 0.052    & 0.667 $\pm$ 0.053    & 0.670 $\pm$ 0.052 & 0.353 $\pm$ 0.102    & 0.630 $\pm$ 0.056    & \underline{0.737 $\pm$ 0.066} & 0.552 $\pm$ 0.116    & \underline{0.788 $\pm$ 0.077} \\
MLP            & 0.697 $\pm$ 0.043    & 0.695 $\pm$ 0.044    & 0.697 $\pm$ 0.043 & 0.396 $\pm$ 0.085    & 0.673 $\pm$ 0.050    & 0.726 $\pm$ 0.043    & 0.653 $\pm$ 0.084    & 0.740 $\pm$ 0.050 \\
RF             & 0.704 $\pm$ 0.055    & \underline{0.703 $\pm$ 0.056} & \underline{0.704 $\pm$ 0.055} & \underline{0.415 $\pm$ 0.109}    & \textbf{0.686 $\pm$ 0.071} & 0.735 $\pm$ 0.058    & 0.664 $\pm$ 0.118 & 0.745 $\pm$ 0.083 \\
Ridge          & 0.650 $\pm$ 0.049    & 0.649 $\pm$ 0.049    & 0.650 $\pm$ 0.049 & 0.303 $\pm$ 0.098    & 0.634 $\pm$ 0.048    & 0.671 $\pm$ 0.059    & 0.631 $\pm$ 0.073    & 0.670 $\pm$ 0.078 \\
XGBoost        & 0.679 $\pm$ 0.078    & 0.678 $\pm$ 0.078    & 0.679 $\pm$ 0.078 & 0.365 $\pm$ 0.161    & 0.657 $\pm$ 0.076    & 0.715 $\pm$ 0.108    & 0.638 $\pm$ 0.107    & 0.721 $\pm$ 0.114 \\
mRMR-mv        & 0.687 $\pm$ 0.051    & 0.686 $\pm$ 0.053    & 0.687 $\pm$ 0.051 & 0.380 $\pm$ 0.099    & 0.670 $\pm$ 0.069    & 0.716 $\pm$ 0.047    & 0.645 $\pm$ 0.126    & 0.729 $\pm$ 0.072 \\
MAgentOmics    & 0.685 $\pm$ 0.035    & 0.684 $\pm$ 0.035    & 0.685 $\pm$ 0.035 & 0.373 $\pm$ 0.070    & 0.666 $\pm$ 0.042    & 0.711 $\pm$ 0.048    & 0.653 $\pm$ 0.078    & 0.718 $\pm$ 0.068 \\
MOGONET        & 0.716 $\pm$ 0.050    & 0.674 $\pm$ 0.060    & 0.673 $\pm$ 0.059 & 0.350 $\pm$ 0.120    & 0.674 $\pm$ 0.073    & 0.681 $\pm$ 0.057    & \underline{0.689 $\pm$ 0.102} & 0.657 $\pm$ 0.079 \\
MOSGAT         & 0.742 $\pm$ 0.054	  & 0.682 $\pm$ 0.063	 & 0.682 $\pm$ 0.062 & 0.366 $\pm$ 0.124    & \underline{0.675 $\pm$ 0.071}	& 0.692 $\pm$ 0.060	   & 0.683 $\pm$ 0.094	  & 0.682 $\pm$ 0.061 \\
\color{blue}PEARL & \color{blue}\underline{0.745 $\pm$ 0.071} & \color{blue}0.659 $\pm$ 0.062 & \color{blue}0.658 $\pm$ 0.060 & \color{blue}0.322 $\pm$ 0.126   & \color{blue}0.670 $\pm$ 0.092	& \color{blue}0.658 $\pm$ 0.051	   & \color{blue}\textbf{0.694 $\pm$ 0.116}	  & \color{blue}0.622 $\pm$ 0.063 \\
HeteroGATomics & \textbf{0.766 $\pm$ 0.046} & \textbf{0.711 $\pm$ 0.042} & \textbf{0.713 $\pm$ 0.041} & \textbf{0.442 $\pm$ 0.076} & \underline{0.675 $\pm$ 0.054} & \textbf{0.783 $\pm$ 0.067} & 0.619 $\pm$ 0.120    & \textbf{0.808 $\pm$ 0.088} \\
\hline
\end{tabular}
\caption{Classification performance comparison (mean $\pm$ standard deviation over 10-fold cross-validation) for the LGG dataset. The best performance for each metric is in bold; second-best results are underlined.}
\label{tbl:class_all_lgg}
\end{table}

\begin{table}[!t]
\centering
\scriptsize
\setlength{\tabcolsep}{1.5mm}
\begin{tabular}{lllllll}
\hline
Method         & Accuracy            & Macro F1            & Micro F1            & Weighted F1         & Precision           & Recall              \\ \hline
KNN            & 0.946 $\pm$ 0.028    & 0.944 $\pm$ 0.025    & 0.946 $\pm$ 0.028    & 0.947 $\pm$ 0.028    & 0.949 $\pm$ 0.027    & 0.946 $\pm$ 0.028 \\
MLP            & 0.954 $\pm$ 0.024    & 0.953 $\pm$ 0.024    & 0.954 $\pm$ 0.024    & 0.953 $\pm$ 0.025    & 0.958 $\pm$ 0.021    & 0.954 $\pm$ 0.024 \\
RF             & 0.950 $\pm$ 0.026    & 0.950 $\pm$ 0.020    & 0.950 $\pm$ 0.026    & 0.950 $\pm$ 0.026    & 0.953 $\pm$ 0.025    & 0.950 $\pm$ 0.026 \\
Ridge          & 0.954 $\pm$ 0.024    & 0.949 $\pm$ 0.025    & 0.954 $\pm$ 0.024    & 0.954 $\pm$ 0.024    & 0.956 $\pm$ 0.024    & 0.953 $\pm$ 0.024 \\
XGBoost        & \underline{0.955 $\pm$ 0.022} & \underline{0.955 $\pm$ 0.019} & \underline{0.955 $\pm$ 0.022} & \underline{0.955 $\pm$ 0.022} & \underline{0.959 $\pm$ 0.020} & \underline{0.955 $\pm$ 0.021} \\
mRMR-mv        & 0.954 $\pm$ 0.021    & 0.953 $\pm$ 0.018    & 0.954 $\pm$ 0.021    & 0.954 $\pm$ 0.022    & 0.956 $\pm$ 0.021    & 0.954 $\pm$ 0.021 \\
MAgentOmics    & 0.950 $\pm$ 0.024    & 0.948 $\pm$ 0.021    & 0.950 $\pm$ 0.024    & 0.950 $\pm$ 0.024    & 0.953 $\pm$ 0.023    & 0.950 $\pm$ 0.023 \\
MOGONET        & 0.952 $\pm$ 0.025    & 0.953 $\pm$ 0.022    & 0.952 $\pm$ 0.025    & 0.952 $\pm$ 0.026    & 0.956 $\pm$ 0.023    & 0.952 $\pm$ 0.025 \\
MOSGAT         & 0.950 $\pm$ 0.022	  & 0.950 $\pm$ 0.017	 & 0.950 $\pm$ 0.022	& 0.950 $\pm$ 0.022	   & 0.954 $\pm$ 0.021	  & 0.950 $\pm$ 0.022 \\
\color{blue}PEARL         & \color{blue}0.952 $\pm$ 0.014	  & \color{blue}0.951 $\pm$ 0.016	 & \color{blue}0.952 $\pm$ 0.014	& \color{blue}0.952 $\pm$ 0.014	   & \color{blue}0.955 $\pm$ 0.013	  & \color{blue}0.952 $\pm$ 0.014 \\
HeteroGATomics & \textbf{0.961 $\pm$ 0.019} & \textbf{0.957 $\pm$ 0.026} & \textbf{0.961 $\pm$ 0.019} & \textbf{0.961 $\pm$ 0.019} & \textbf{0.964 $\pm$ 0.019} & \textbf{0.961 $\pm$ 0.019} \\
\hline
\end{tabular}
\caption{Classification performance comparison (mean $\pm$ standard deviation over 10-fold cross-validation) for the RCC dataset. The best performance for each metric is in bold; second-best results are underlined.}
\label{tbl:class_all_rcc}
\end{table}

{\color{blue}
\subsection{Computational Cost}
We further analyse the computational cost of HeteroGATomics, including its time complexity, runtime, scalability, and memory usage (Appendix~D). The complexity analysis shows that the cost of HeteroGATomics\textsubscript{MAS} is dominated by the iterative multi-agent search, whereas the cost of HeteroGATomics\textsubscript{GAT} increases as additional omics modalities are incorporated (Appendix~D.1). In practice, HeteroGATomics\textsubscript{MAS} requires approximately 1.2--2.5 hours but is run once for each training set as a preprocessing step and can be parallelized across agents (Appendix~D.2). Notably, it is considerably faster than both multimodal feature selection baselines, mRMR-mv and MAgentOmics, while achieving higher predictive performance. HeteroGATomics\textsubscript{GAT} requires approximately 63--84 seconds, a moderate increase over other graph-based methods such as PEARL, MOSGAT, and MOGONET, while achieving the best overall predictive performance (Appendix~D.3). Moreover, HeteroGATomics\textsubscript{GAT} contains approximately 1.13--1.50 million trainable parameters and requires approximately 0.9--1.2 GB of peak GPU memory, while HeteroGATomics\textsubscript{MAS} has no trainable parameters (Appendix~D.4).
}

\subsection{Ablation Studies}
\label{sec:ablation}
\begin{figure}[!t]
\color{blue}
\centering
    \includegraphics[width=0.95\textwidth]{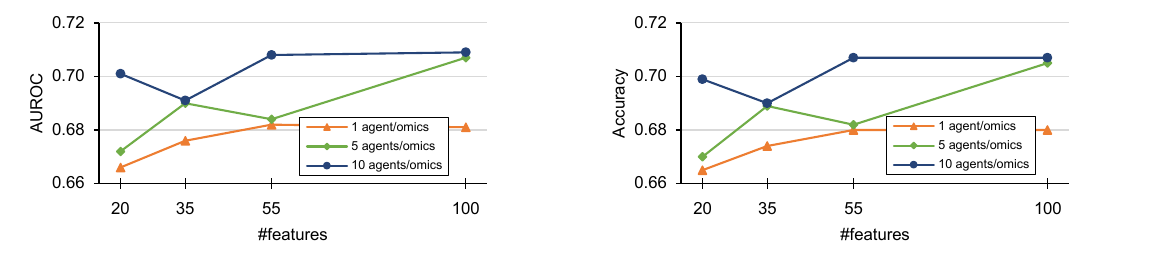}
    \caption{Performance of HeteroGATomics\(_{\mathrm{MAS}}\) with different numbers of agents for random forest classification on the LGG dataset. The averaged values from 10-fold cross-validation are reported for each metric. Results are reported for 1, 5, and 10 agents per omics modality.}
    \label{fig:feat_agent}
\end{figure}

We conduct ablation studies to examine key design choices and individual modules within HeteroGATomics and assess its effectiveness in integrating different modalities. {\color{blue}First, we investigate the contribution of the multi-agent design within HeteroGATomics\textsubscript{MAS} by varying the number of agents per omics modality. Figure~\ref{fig:feat_agent} compares configurations with 1, 5, and 10 agents on LGG across different selected feature sizes. Performance generally improves when moving from a single-agent to a multi-agent configuration, while the smaller differences between 5 and 10 agents at some feature sizes suggest diminishing returns as the number of agents increases. These results support the use of multiple agents for exploring the multimodal feature space.}

\begin{figure}[!t]
\centering
    \includegraphics[width=0.9\textwidth]{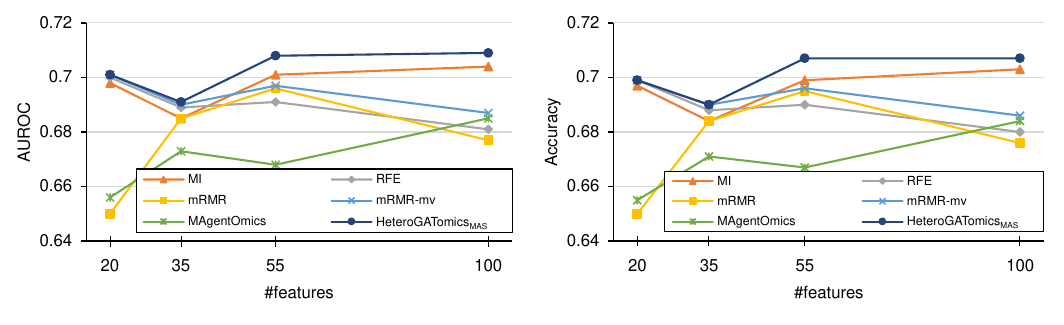}
    \caption{Performance comparison of feature selection methods for random forest classification on LGG. The averaged values from 10-fold cross-validation are reported for each metric.}
    \label{fig:feat_sizes}
\end{figure}

We then investigate the effectiveness of the standalone HeteroGATomics\textsubscript{MAS} module. Figure~\ref{fig:feat_sizes} presents the classification results of different feature selection methods on LGG using RF, with logarithmically spaced selected feature sizes. HeteroGATomics\textsubscript{MAS} has consistently outperformed other methods in terms of AUROC and accuracy. MI, a univariate method, achieves the second-highest performance, even against multivariate methods. This suggests that LGG may contain individual features with significant independent predictive power, and the complex interactions identified by multivariate methods may not significantly enhance performance. This finding highlights the effectiveness of HeteroGATomics\textsubscript{MAS} in identifying discriminative features despite its multivariate nature. Interestingly, both AUROC and accuracy metrics yield remarkably similar trends across all methods, indicating a well-balanced trade-off between true and false positive rates.

\begin{figure}[!t]
\centering
    \includegraphics[width=0.8\textwidth]{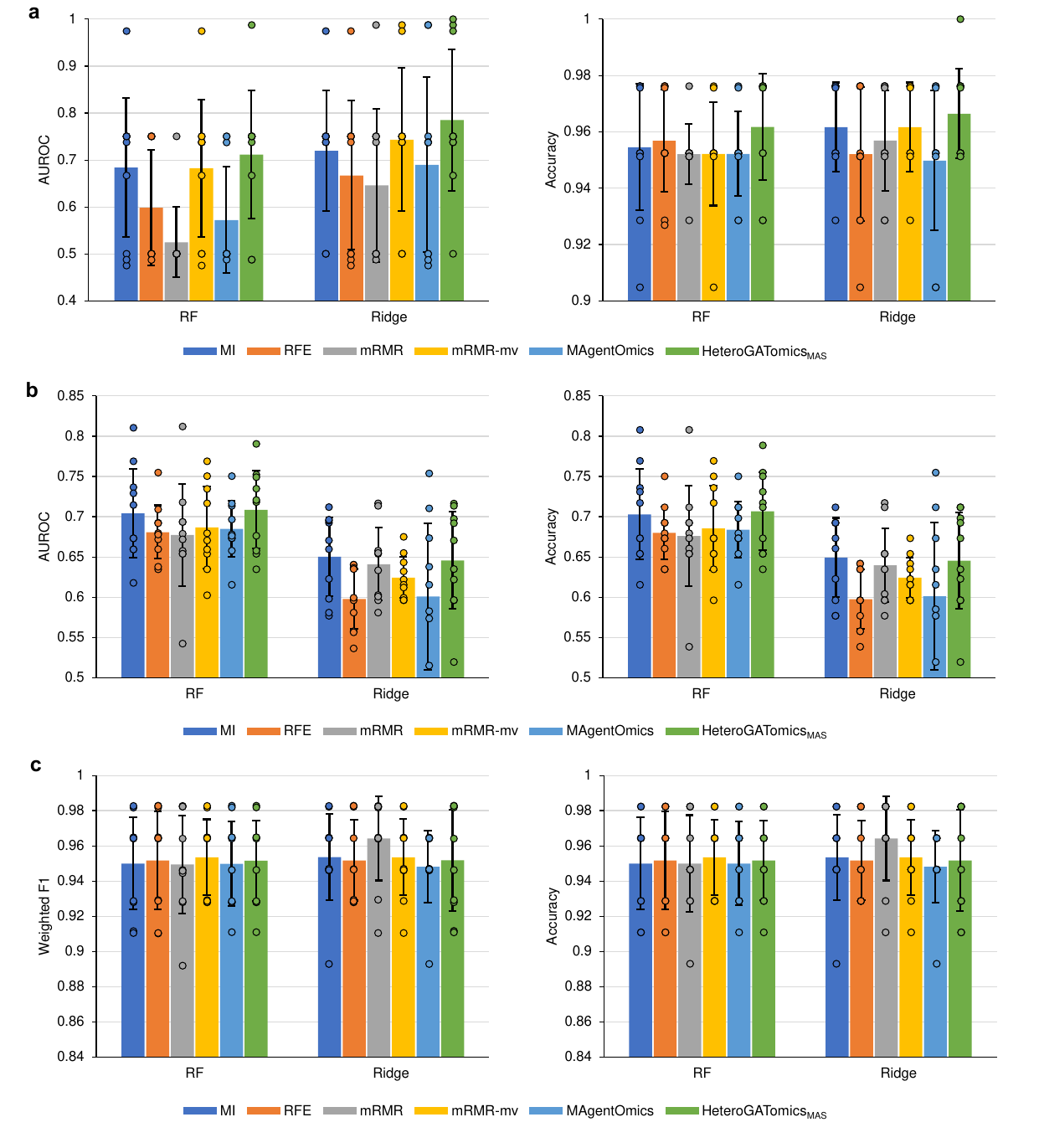}
    \caption{Performance comparison of feature selection methods for random forest and Ridge classification for (a) the BLCA, (b) the LGG, and (c) the RCC datasets. The vertical bars show the mean, the black lines represent error bars indicating plus/minus one standard deviation, and each dot is a model's performance on each fold.}
\label{fig:feat_classifiers}
\end{figure}

Figure~\ref{fig:feat_classifiers} further validates the generalizability of HeteroGATomics\textsubscript{MAS} by comparing its performance across three datasets. The figure compares results for 100 selected features from each modality, evaluated with two classifiers, RF and Ridge. On BLCA, HeteroGATomics\textsubscript{MAS} outperforms all baselines across both evaluation criteria with both classifiers, surpassing the best-reported baseline by a relative improvement of 4.0\% with RF and 5.6\% with Ridge in AUROC (Fig.~\ref{fig:feat_classifiers}(a)). For LGG, HeteroGATomics\textsubscript{MAS} continues to excel in identifying high-quality features, leading to superior evaluation metrics with RF (Fig.~\ref{fig:feat_classifiers}(b)). While MI slightly outperforms HeteroGATomics\textsubscript{MAS} by less than 0.7\% in AUROC and accuracy with Ridge, HeteroGATomics\textsubscript{MAS} remains the top performer in both metrics when using RF, even surpassing Ridge's results. RCC, known for its relatively straightforward nature in the classification task \cite{wang2021mogonet}, sees all methods achieving high performance in both metrics with both classifiers (Fig.~\ref{fig:feat_classifiers}(c)). HeteroGATomics\textsubscript{MAS} remains competitive on RCC when using RF, with mRMR only outperforming it under Ridge. Additionally, we observe that the efficacy of a given feature selection technique is highly dependent on the dataset. mRMR, for example, is the superior method for RCC, yet it significantly underperforms compared to baselines for BLCA when Ridge is used. 

{\color{blue}The performance variation across downstream classifiers in Fig.~\ref{fig:feat_classifiers} is primarily dataset-dependent: all feature selection methods achieve higher AUROC with Ridge on BLCA and with RF on LGG, while both classifiers perform comparably on RCC. This suggests that classifier preference reflects dataset characteristics; for example, the class imbalance in BLCA may favour regularised linear models, whereas the balanced LGG cohort may benefit from the ability of RF to capture nonlinear feature interactions. Importantly, the relative performance of HeteroGATomics\textsubscript{MAS} compared with the baselines is largely preserved across classifiers, indicating that the selected multimodal features remain effective across different downstream classifier.} These analyses highlight HeteroGATomics\textsubscript{MAS}'s ability to effectively identify discriminative features across different datasets and classifiers. 

\begin{figure}[!t]
\centering
    \includegraphics[width=0.98\textwidth]{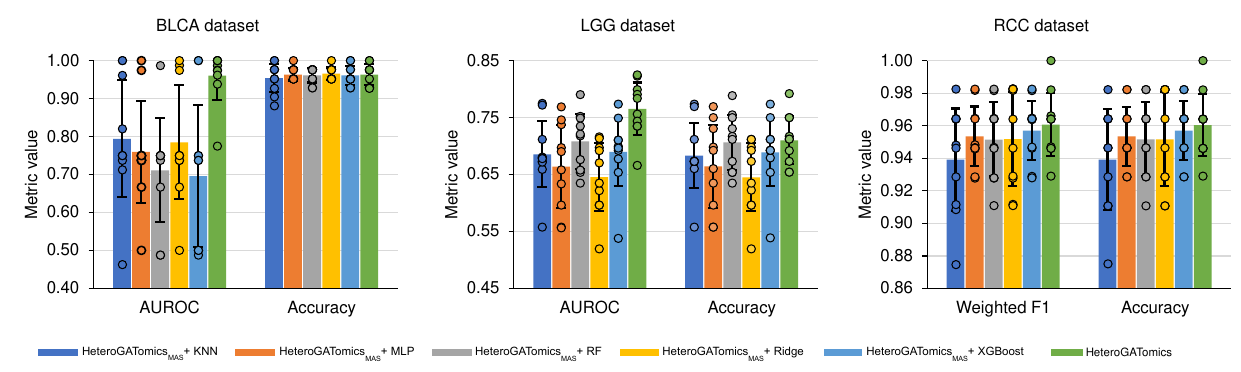}
    \caption{Ablation studies to evaluate the feature selection module across five classifiers. HeteroGATomics\textsubscript{MAS} + [classifier] denotes the results of the feature selection module within HeteroGATomics for a classifier, while HeteroGATomics represents the results derived from the entire HeteroGATomics architecture.}
\label{fig:ablation_feat}
\end{figure}

Next, we highlight the crucial role of HeteroGATomics\textsubscript{GAT} in augmenting the predictive capabilities of HeteroGATomics architecture for classification tasks, beyond what its HeteroGATomics\textsubscript{MAS} can achieve. Figure~\ref{fig:ablation_feat} compares the whole HeteroGATomics pipeline against HeteroGATomics\textsubscript{MAS} on BLCA, LGG, and RCC. We show HeteroGATomics\textsubscript{MAS}'s performance using KNN, MLP, RF, Ridge, and XGBoost classifiers. Figure~\ref{fig:ablation_feat} shows the addition of the heterogeneous GAT module (HeteroGATomics\textsubscript{GAT}) boosts HeteroGATomics's performance by 20.9\% and 8.1\% in AUROC on BLCA and LGG, respectively, and by 0.4\% in weighted F1 on RCC, compared with the best-performing classifiers using features selected by HeteroGATomics\textsubscript{MAS}. These results highlight the value of the heterogeneous GAT module in HeteroGATomics to enhance its multiomics integration capabilities.

\begin{figure}[!t]
\centering
\includegraphics[width=0.6\textwidth]{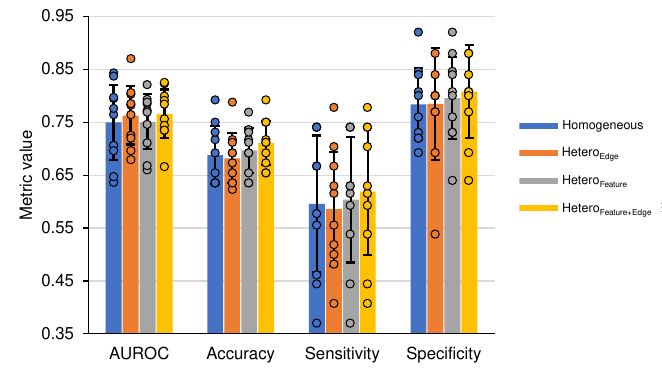}
\caption{Ablation studies to evaluate heterogeneous graph elements using LGG.}
\label{fig:ablation_hetero_graph}
\end{figure}

Then, we demonstrate the effectiveness of heterogeneous graphs and their impact on GAT performance. We conduct several modifications: removing the feature similarity network to simulate a homogeneous graph scenario (denoted as Homogeneous), removing static and dynamic edge weights derived from the MAS module as edge attributes (denoted as Hetero\textsubscript{Feature}), and excluding static and dynamic node scores derived from the same module as node attributes (denoted as Hetero\textsubscript{Edge}). We refer to the full HeteroGATomics architecture, incorporating both node and edge attributes, as Hetero\textsubscript{Feature+Edge}. The results of this comparative study on LGG in Fig.~\ref{fig:ablation_hetero_graph} show that HeteroGATomics, when leveraging heterogeneous graphs, outperforms its homogeneous graph counterpart, achieving improvements of 3.3\%, 2.1\%, 3.9\%, and 3.1\% in accuracy, AUROC, sensitivity, and specificity, respectively. Furthermore, the enhancement in HeteroGATomics' performance comes not only from the feature similarity network but also from the contributions of each individual network element, including feature and edge attributes.

\begin{figure}[!t]
\centering
\color{blue}
\includegraphics[width=0.8\textwidth]{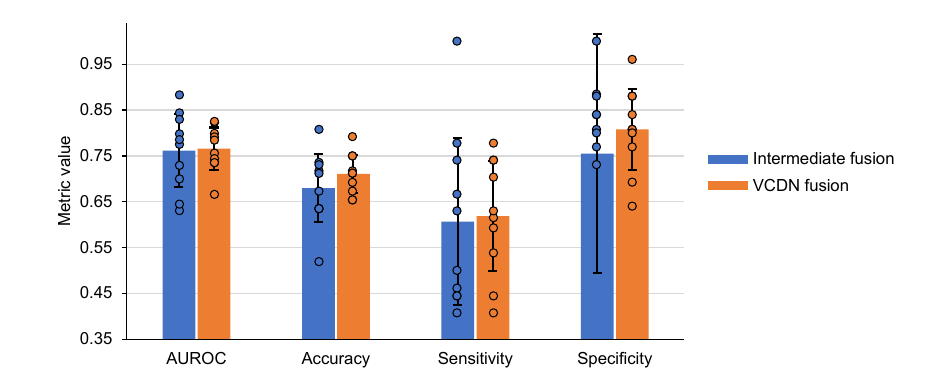}
\caption{Ablation studies to evaluate the impact of the fusion strategy on HeteroGATomics using LGG.}
\label{fig:ablation_vcdn}
\end{figure}

{\color{blue}Moreover, to evaluate the contribution of VCDN to multimodal integration, we compare VCDN fusion with an intermediate fusion variant, in which the representations learned by the omics-specific encoders are concatenated for final classification, while keeping the remaining architecture unchanged. As shown in Fig.~\ref{fig:ablation_vcdn}, VCDN improves AUROC, accuracy, sensitivity, and specificity by approximately 0.5\%, 4.6\%, 2\%, and 7\%, respectively. These results demonstrate that VCDN provides measurable improvements over intermediate fusion, with the largest gain in specificity, suggesting that it better exploits complementary evidence across omics modalities to distinguish the two classes.}

\begin{figure}[!t]
\centering
\includegraphics[width=0.7\textwidth]{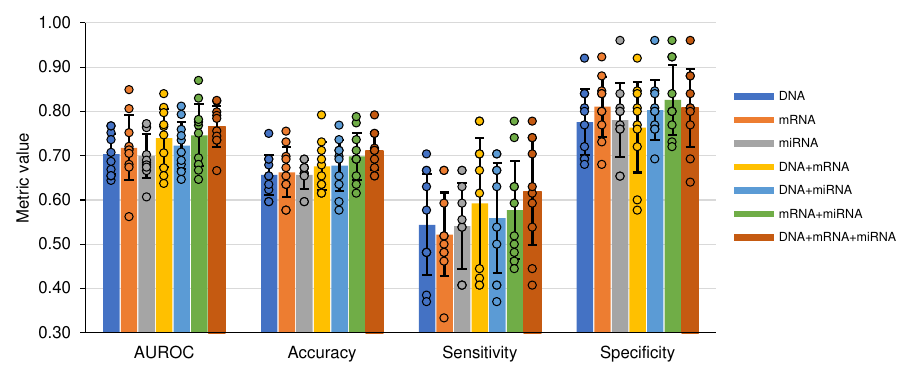}
\caption{Ablation studies to evaluate the impact of different omics combinations on HeteroGATomics performance using LGG.}
\label{fig:ablation_modality}
\end{figure}

To evaluate the impact of different omics modalities on HeteroGATomics' performance and to determine the benefit of integrating multiple omics, we train HeteroGATomics using seven modality combinations: individual modalities without the VCDN module (DNA, mRNA, and miRNA), combinations of two modalities (DNA+mRNA, DNA+miRNA, mRNA+miRNA), and all three modalities together (DNA+mRNA+miRNA). The performance of these configurations is assessed on LGG in AUROC, accuracy, sensitivity, and specificity. The results in Fig.~\ref{fig:ablation_modality} show that integrating additional omics modalities with HeteroGATomics enhances its performance across various evaluation metrics, except for specificity. In the context of single-modality learning, mRNA contributes significantly to the performance, particularly in AUROC and accuracy. Notably, integrating all three modalities outperforms single or dual-modality combinations, further highlighting the benefits of multiomics integration via HeteroGATomics.

{\color{blue}
\subsection{Quantified Contribution of Design Choices}
Table~\ref{tbl:ablation_summary} presents selected results from the ablation studies in Section~\ref{sec:ablation}, linking each major design choice in HeteroGATomics to the challenge it addresses and quantifying its contribution to performance through experiments A1 to A7. The multi-agent design (A1) addresses the large combinatorial search space of multimodal feature selection by enabling multiple agents to explore diverse candidate feature subsets within and across omics modalities. Through shared dynamic node scores and edge weights, information from previous searches guides subsequent exploration, balancing exploration and exploitation. Using ten agents instead of one improves AUROC by 2.2--5.3\% and accuracy by 2.4--5.1\% across all evaluated feature sizes, supporting this design choice. HeteroGATomics\textsubscript{MAS} (A2) is used as the feature selection module because it identifies informative features, with AUROC gains of 4.0--5.6\% on BLCA and comparable performance on LGG and RCC, while also generating node and edge information used for downstream graph construction.

The complete HeteroGATomics framework improves upon the standalone feature selection module (A3), with AUROC gains of 20.9\% on BLCA and 8.1\% on LGG and a smaller gain of 0.4\% in Weighted F1 on RCC. These results show that the additional heterogeneous graph learning and multimodal integration stages provide measurable performance improvements. Within the graph learning stage, the use of heterogeneous graphs (A4) and the retention of both node and edge information (A5) are motivated by the need to represent multiple levels of relationships in multiomics data. Heterogeneous graphs model patient--patient, feature--feature, and feature--patient relations simultaneously, whereas the homogeneous variant captures only patient--patient relations. This design improves all four evaluated metrics by 2.1--3.9\%. Retaining both node and edge information further improves accuracy, sensitivity, and specificity by 1.5--2.5\%, with a smaller gain of 0.4\% in AUROC, indicating that the information generated by HeteroGATomics\textsubscript{MAS} is beneficial beyond feature selection itself.

For multimodal fusion, VCDN (A6) is used to capture cross-omics interactions among the omics-specific predictions rather than simply concatenating intermediate representations. Compared with intermediate fusion, VCDN improves AUROC, accuracy, sensitivity, and specificity by 0.5\%, 4.6\%, 2.0\%, and 7.0\%, respectively, supporting this design choice.

Finally, integrating all three omics modalities (A7) improves AUROC by 6.7\% over the best one-omics configuration and by 2.6\% over the best two-omics configuration, supporting the contribution of complementary information across omics modalities.

\begin{table}[!t]
\color{blue}
\centering
\scriptsize
\begin{tabular}{p{0.5cm}p{4.5cm}p{6.0cm}p{3.6cm}}
\hline
ID & Design question & Ablation study & Key performance change \\
\hline
A1 & Why use multiple agents per omics modality instead of a single agent? & 1 vs 10 agents per omics modality (Fig.~\ref{fig:feat_agent}) \newline RF classifier on LGG dataset \newline 20--100 selected features  & +2.2--5.3\% AUROC \newline +2.4--5.1\% Accuracy \\
\hline
A2 & Why use HeteroGATomics\textsubscript{MAS} rather than existing unimodal and multimodal feature selection methods? & HeteroGATomics\textsubscript{MAS} vs best feature selection baseline (Fig.~\ref{fig:feat_classifiers}) \newline 100 features selected per omics modality & BLCA: \newline +4.0\% AUROC (RF) \newline +5.6\% AUROC (Ridge) \newline LGG: \newline +0.7\% AUROC (RF) \newline $-$0.7\% AUROC (Ridge) \newline RCC: \newline $-$0.2\% Weighted F1 (RF) \newline $-$1.3\% Weighted F1 (Ridge) \\
\hline
A3 & Why add graph learning and multimodal fusion on top of feature selection? & Full HeteroGATomics vs HeteroGATomics\textsubscript{MAS} with the best classifier (Fig.~\ref{fig:ablation_feat}) & +20.9\% AUROC (BLCA) \newline +8.1\% AUROC (LGG) \newline +0.4\% Weighted F1 (RCC) \\
\hline
A4 & Why use heterogeneous rather than homogeneous graphs? & Heterogeneous (patient and feature similarity networks) vs homogeneous (patient similarity network) graphs (Fig.~\ref{fig:ablation_hetero_graph}) \newline LGG dataset & +2.1\% AUROC \newline +3.3\% Accuracy \newline +3.9\% Sensitivity \newline +3.1\% Specificity \\
\hline
A5 & Why retain node and edge information from the HeteroGATomics\textsubscript{MAS} module? & Both node and edge information vs only one of them (best variant) (Fig.~\ref{fig:ablation_hetero_graph}) \newline LGG dataset & +0.4\% AUROC \newline +2.0\% Accuracy \newline +2.5\% Sensitivity \newline +1.5\% Specificity \\
\hline
A6 & Why use VCDN for multimodal fusion? & VCDN vs intermediate fusion (concatenation of embeddings) (Fig.~\ref{fig:ablation_vcdn}) \newline LGG dataset
& +0.5\% AUROC \newline +4.6\% Accuracy \newline +2.0\% Sensitivity \newline +7.0\% Specificity \\
\hline
A7 & Why integrate all three omics modalities? & All three omics vs best one- and two-omics configurations (Fig.~\ref{fig:ablation_modality}) \newline LGG dataset & vs best one-omics: \newline +6.7\% AUROC \newline +7.2\% Accuracy \newline vs best two-omics: \newline +2.6\% AUROC \newline +1.7\% Accuracy \\
\hline
\end{tabular}
\caption{Selected ablation results and quantified contributions of key HeteroGATomics design choices. Values are relative changes in mean performance over 10-fold cross-validation. Positive values indicate improvements of the proposed design over the corresponding ablated or simplified alternative.}
\label{tbl:ablation_summary}
\end{table}
}

\subsection{Analysis of Identified Biomarkers for Cancer Diagnosis}
To understand the decision-making process within HeteroGATomics, we utilize a biomarker extraction technique (see Appendix F) to identify and analyze the key biomarkers that significantly influence the classification results. This process is essential for interpreting the architecture's effectiveness and pinpointing the most informative features serving as cancer biomarkers. Supplementary Tables G.5 and G.6 present the 30 most important biomarkers identified by HeteroGATomics on BLCA and LGG, respectively. {\color{blue}Since biomarker importance is evaluated on the features selected by HeteroGATomics\textsubscript{MAS} in each fold, we further assess the stability of these candidate biomarkers across folds. Of the 300 features selected per fold, 175 (58\%) on LGG and 179 (60\%) on BLCA are consistently selected in all 10 folds, indicating a stable core set of candidate biomarkers across different training sets (Appendix~G).} RCC is excluded from further biomarker identification, serving only as a proof-of-concept for multi-class classification tasks \cite{wang2021mogonet}.

Top ranking biomarkers for LGG consist of 16 DNA methylation features, 13 mRNAs, and 1 miRNA. BLCA consists of 14 mRNA features, 11 DNA methylation features, and 5 miRNAs. Gene ontology (GO) enrichment analysis on the top 30 DNA and mRNA biomarkers highlights terms related to synaptic and signal transduction for LGG (GO:0097060, $p$=0.009 and GO:0007165, $p$=0.012, see Supplementary Fig. G.7(a) in Appendix G). Belonging to those categories is PTPRA, a protein tyrosine phosphatases with known roles in tumorigenesis \cite{lv2023TheRO}. CHRND and KCNC2, ion channels genes, also belong to both categories. Numerous ion channels are dysregulated in glioma and significantly impact prognosis. Xu \textit{et al.} have previously identified KCNC2 as downregulated in malignant gliomas \cite{xu2017screening}, while CHRND has been identified as a biomarker for treatment and prognosis of head and neck cell carcinomas \cite{li2021identification}. 

GO terms related to the top BLCA biomarkers include several developmental processes (e.g. Regionalization, $p$=0.008, see Supplementary Fig. G.7(b) in Appendix G). This emphasises top biomarkers that are Transcription Factors (TFs), essential regulators of gene expression: YBX2, HOXB2 and 3, FOXH1, FOXA3, DMRTA2, TEX15 and MAGEA10. These have diverse functions: DNA repair, cell differentiation and migration, cell cycle and organogenesis to cite a few, all relevant in the context of cancer. Among them, TFs of the Fox superfamily FOXH1 and FOXA3 are identified as mRNA biomarkers. FOX family proteins have been shown to be involved in bladder development and cancer progression, for instance FOXA1 expression has been correlated to poor survival. For DNA methylation features, HOXB2 has recently been shown to be upregulated in some subtypes of BLCA and to act as a tumor promoter \cite{liu2019hoxb2}. All of the following genes identified by HeteroGATomics have been previously flagged as potential biomarkers for BLCA: YBX2 \cite{yuan2024comprehensive}, DMRTA2 \cite{deng2022novel}, TEX15 \cite{mantere2017case} and MAGEA10  \cite{verma2024melanoma}.

Notably, the top 5 biomarkers of BLCA are all miRNAs, small RNAs known for regulating gene expression post-transcriptionally by binding target mRNAs and preventing their translation via degradation or translational silencing. As such they play a central role in cell maintenance and in tumorigenesis. Numerous studies have pointed out the importance of miRNAs in bladder cancer and their potential use in cancer diagnosis and prognosis \cite{das2023roles}. All 5 miRNAs have been studied in the context of other cancers, for example pancreas and liver \cite{lee2023anti} or breast \cite{khodadadi2018prognostic}. Three of our five (hsa-mir-1-3p, hsa-mir-708-5p and hsa-mir-16-2-3p), have been previously linked to bladder cancer \cite{tan2022dysregulation}.  Zhang \textit{et al.} demonstrated that hsa-mir-1-3p is downregulated in bladder cancer tissues and is able to inhibit cancer proliferation and tumorigenesis in vivo \cite{zhang2018mir}. HeteroGATomics is able to not only identify known miRNAs related to bladder cancer but also identify new potential miRNAs, hsa-mir-1976 and hsa-mir-24-3p, involved in oncogenesis.

\begin{figure}[!t]
\centering
    \includegraphics[width=0.95\textwidth]{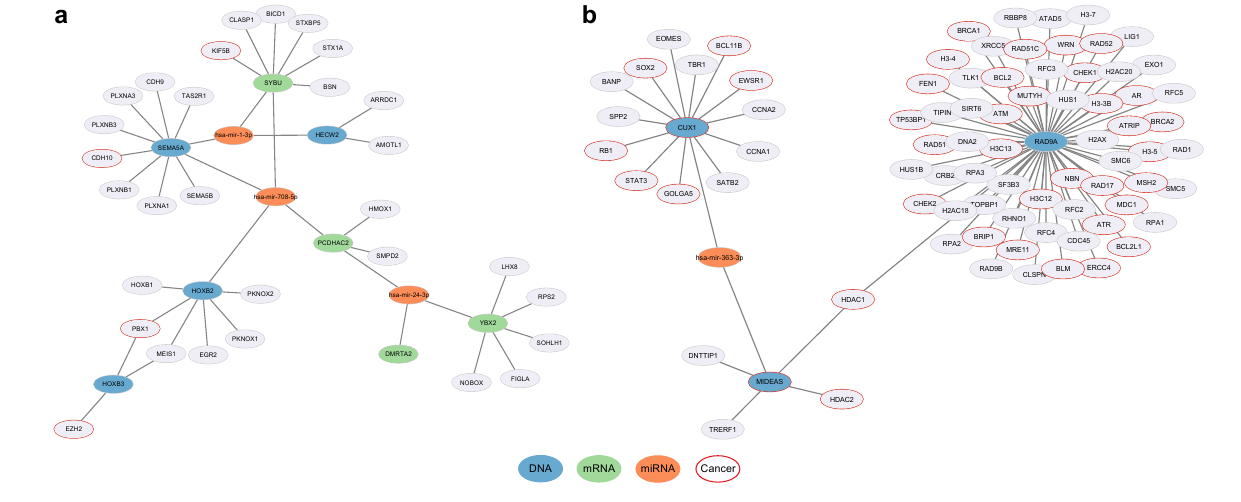}
    \caption{Known partners of selected top biomarkers for (a) the BLCA dataset and (b) the LGG dataset. Direct protein-protein interactions are recovered for DNA and mRNA omics. For the miRNA omics, known mRNA targets are recovered from starBase.}
    \label{fig:partners_selected}
\end{figure}

We further investigate the interaction network of the top biomarkers, including protein-binding partners of protein coding biomarkers and targets of miRNAs biomarkers (Fig.~\ref{fig:partners_selected} and Supplementary Figs. G.5 and G.6 in Appendix G). Interestingly, the BLCA miRNA biomarker hsa-mir-708-5p targets four of the protein-coding biomarkers, while hsa-mir-1-3p targets three. This includes partners such as HOXB2 and YBX2, suggesting a possible regulation between them (Fig.~\ref{fig:partners_selected}(a)). Examining the protein-protein interactions of the LGG biomarkers shows that RAD9A has numerous protein partners and many are known genes related to cancer (Fig.~\ref{fig:partners_selected}(b) and Supplementary Fig. G.6 in Appendix G). RAD9A itself is involved in DNA repair, interacts with several components of the DNA damage response pathways that is targeted for glioblastoma treatment \cite{ferri2020targeting}. For instance, RAD9A interacts with ATR to mediate DNA repair and several ATR inhibitor drugs have been developed for treating Glioblastoma \cite{ferri2020targeting}. RAD9A also interacts with HDAC1, a mediator of chromatin compaction, known to be frequently overexpressed in LGG and also targeted for therapy \cite{cascio2021nonredundant}. Interestingly, another biomarker, MIDEAS, also interacts with HDAC1, as well as with HDAC2 (Fig.~\ref{fig:partners_selected}(b)), forming part of the mitotic deacetylase (MiDAC) complex, which is important for neuronal development \cite{mondal2020histone}. 

The homeobox TF CUX1 is another important component of LGG. Like MIDEAS, CUX1 is also a target of the miRNA biomarker hsa-mir-363-3p (Fig.~\ref{fig:partners_selected}(b)). Moreover, CUX1 has been previously identified as widely expressed in glioma. CUX1 seems to promote tumorigenesis via the Wnt/b-Catenin pathway. Another biomarker, RBMS3, is an RNA binding protein that regulates crucial cellular processes such as transcription, cell apoptosis or cell cycle progression. Its expression correlates with good or poor prognosis depending on cancer type. Ruan \textit{et al.} recently discovered that RBMS3 downregulation leads to increased proliferation of glioblastoma cells by interacting with circHECTD1. Subsequently increasing VE-cadherin levels and promoting vasculogenic mimicry, which has a deleterious effect on anti-vascular therapy \cite{ruan2023rbms3}. HeteroGATomics also highlights deletion of BRINP1, also known as DBC1 for deleted in bladder cancer 1. Silencing of, or deletions on, chromosome 9 are known in the context of bladder cancer and include silencing of  BRINP1.  BRINP1 has been shown to suppress cell cycle progression and to be a candidate tumor supressor \cite{nishiyama2001negative}. 

\section{Discussion}
The development of cancer involves multiple biological layers. The availability of diverse biological data, multiomics, offers detailed insights into various cancer mechanisms. However, integrating multiple heterogeneous omics datasets for cancer diagnosis and biomarker identification remains a significant challenge, especially in small patient cohorts with high-dimensional feature spaces. Our HeteroGATomics is a GAT-based method enhanced with heterogeneous graphs to integrate multiomics data for cancer diagnosis. To construct these heterogeneous graphs, we employ a novel dual-view representation leveraging feature and patient similarity networks. HeteroGATomics utilizes a multimodal feature selection approach using a multi-agent system, effectively addressing the high dimensionality of the feature space.

Experimental results demonstrate that the standalone multimodal feature selection module of HeteroGATomics {\color{red}outperforms baseline feature selection methods in most cases and remains competitive in the remaining cases, with advantages observed on BLCA and LGG, while all methods perform comparably on RCC.} This suggests that employing an effective multimodal feature selection strategy across multiple modalities can outperform independent feature selection for each modality. Another key innovation of HeteroGATomics is the construction of heterogeneous graphs through our proposed dual-view representation. {\color{red}Leveraging these heterogeneous graphs enhances the performance of HeteroGATomics over the standalone feature selection module across all datasets, with substantial AUROC gains on BLCA and LGG. Furthermore, experiments show that HeteroGATomics outperforms both conventional and state-of-the-art multiomics integration methods in cancer classification tasks across most evaluation metrics, with the remaining cases reflecting sensitivity-specificity trade-offs.} {\color{blue}External validation further supports the generalizability of HeteroGATomics. When trained on TCGA-BRCA and tested on the independent METABRIC cohort, it outperforms graph-based multiomics integration methods across nearly all evaluation metrics and remains stable across feature sizes, despite differences in measurement platforms between the two cohorts.}

Ablation studies demonstrate that each component of the heterogeneous graphs positively impacts HeteroGATomics' performance compared to homogeneous graphs. This highlights the importance of leveraging auxiliary information inherent in multiomics datasets for cancer classification and {\color{red}the effectiveness of HeteroGATomics in utilizing this information.} Additionally, ablation studies reveal the advantage of training HeteroGATomics with multiple omics over fewer omics, reflecting the complementary information each omic provides to the model. HeteroGATomics is flexible, enabling the integration of additional omics that contribute to understanding cancer progression.

Interpretability is another key aspect of this study, essential for building trust in HeteroGATomics. We enhance interpretability through a biomarker identification process that explains which features have the most significant impact on the cancer classification. HeteroGATomics successfully predicts known biomarkers in each cohort, including known genes involved in tumorigenesis, diagnosis and/or treatment. We show via interaction networks that several biomarkers are associated and highlight important functions that are or could be targeted for therapy (e.g. DNA damage repair pathways in LGG). HeteroGATomics also identifies genes that were not previously studied in LGG or BLCA, with several known oncogenes in other cancers. Examples of novel therapy target candidates for LGG include ion channels genes CHRND and KCNC2. For BLCA, the top five miRNAs biomarkers can progress miRNA pools used for diagnosis and prognosis. We highlight two network clusters of either interconnected markers, or markers that share common interactors. These may well be highlighting key regulatory processes, where a single process can be disrupted in multiple ways to achieve oncogenicity. HeteroGATomics also uncovers non protein coding genes of interest like pseudogenes including long non coding RNAs (lncRNAs, e.g. RNF126P1 and TTTY14 in LGG). Their relevance in cancer has been more recently investigated. For instance dysregulation of lncRNAs has been shown to impact glioma development. 

{\color{blue}From a clinical perspective, HeteroGATomics could support cancer diagnosis by integrating multiple molecular data types into a single predictive framework and by highlighting candidate biomarkers that contribute to the prediction. Such information may support diagnostic classification when multiple omics modalities are available. The biomarker identification component may also assist in generating hypotheses for further biological and translational investigation. However, validation across additional cancer types, sequencing platforms, and clinical settings will be important before clinical translation.}

{\color{blue}As is common in cancer multiomics studies, the available cohorts contain relatively small numbers of patients compared with the high dimensionality of the molecular data. Although HeteroGATomics is designed to address this setting, and its performance is supported by both 10-fold cross-validation and external validation on METABRIC, the external validation is limited to a single cancer type. Extending such validation to other cancer types would provide further evidence of its generalizability.}

\section{Conclusion}
In this work, we present HeteroGATomics, a novel deep learning framework based on GATs for integrating multiomics data in cancer diagnosis and biomarker identification. Our model incorporates an MAS for multimodal feature selection, which also provides auxiliary structural information derived from the selected features. Furthermore, we introduce a dual-view representation for constructing heterogeneous graphs that capture patient-level and feature-level structures, enabling a holistic approach to graph representation learning. {\color{red}Comprehensive experiments on three public cancer multiomics datasets show that HeteroGATomics outperforms baseline methods across most evaluation metrics.} {\color{blue}External validation, training on TCGA-BRCA and testing on the independent METABRIC cohort, further shows that HeteroGATomics generalizes across cohorts and measurement platforms, outperforming graph-based integration methods.} Moreover, ablation studies validate the contribution of each module, and our biomarker identification process highlights the model’s effectiveness in discovering cancer-related biomarkers.

Future work can expand the flexible architecture of HeteroGATomics to incorporate additional modalities beyond biological omics, such as medical imaging and electronic health records. This extension would allow the model to learn more complex relationships across multiple modalities. Moreover, the dual-view representation concept in HeteroGATomics is broadly applicable, extending beyond the specific problem and datasets used in this study. This approach can be adapted to tabular-structured datasets and applied across various fields beyond cancer genomics, wherever explicit/implicit meaningful relationships among features exist. HeteroGATomics, similar to many studies, requires patients to have all omics available and cannot handle patients with one or more omics missing. Enabling HeteroGATomics to learn from missing omics will improve its performance in multiomics datasets.

\section*{Declaration of competing interests}
The authors declare that they have no known competing financial interests or personal relationships that could have appeared to influence the work reported in this paper.

\section*{Acknowledgements}
This work was supported by the University of Sheffield Faculty of Engineering Research Scholarship (grant number 199256787) and the Medical Research Council (MRC) grant MR/V010948/1.

\bibliographystyle{elsarticle-num} 
\bibliography{main}

\newpage

\begin{bibunit}
\addtolength{\topmargin}{-1.5cm}
\addtolength{\textheight}{3cm}

\begin{center}
    \Large Supplementary Material: Heterogeneous Graph Attention Network With Multimodal Feature Selection for Cancer Multiomics Integration \\
    \vspace{1.5em}
    \normalsize Sina Tabakhi, Charlotte Vandermeulen, Ian Sudbery, Haiping Lu
\end{center}
\thispagestyle{pprintTitle}

\appendix
\section{Notations}
\label{app:not}
The important notations used in this paper are provided in Table~\ref{tbl:notation}. In this paper, vectors are denoted by lowercase boldface letters, matrices by uppercase boldface letters, tensors by uppercase bold italic letters, and sets by calligraphic letters.

\begin{table}[h]
\centering
\footnotesize
\caption{Notations and descriptions.\label{tbl:notation}}
\setlength{\tabcolsep}{2pt}
\renewcommand{\arraystretch}{1.3}
{\begin{tabular}{lp{12.8cm}}
\toprule
Notation & Description \\ 
\midrule
${\mathbf{a}}_{r,s}^{(l)}, {\mathbf{a}}_{r,d}^{(l)}, {\mathbf{a}}_{r,e}^{(l)}$ & Attention vectors to weigh source, destination, and edge attributes for relation $r$ in $l$th layer \\
$\mathcal{D}^m=<\mathbf{X}^m,\mathbf{Y}>$ &	Omics $m$ with patients $\mathbf{X}^m\in\mathbb{R}^{N\times d_m}$ and labels $\mathbf{Y}\in\mathbb{R}^{{N \times C}}$ \\
${\mathbf{e}}_{uv}$ & Edge attributes between nodes $u$ and $v$ \\
$\eta_\mathcal{V}(\cdot),\eta_\mathcal{E}(\cdot)$ & Static node score and edge weight \\
$\mathcal{G}^m=(\mathcal{V}^m, \mathcal{E}^m, w_m)$ & Undirected weighted graph with nodes $\mathcal{V}^m$, edges $\mathcal{E}^m$, and omics importance $w_m$ \\
$\mathcal{G}_H^m=(\mathcal{V}^m,\mathcal{E}^m,\mathcal{O},\mathcal{R})$ & Heterogeneous graph with nodes $\mathcal{V}^m$, edges $\mathcal{E}^m$, and corresponding types $\mathcal{O}$, $\mathcal{R}$ \\
$\mathbf{h}_u^{(l)}$ & Embedding of node $u$ in the $l$th layer\\
$\mathcal{J}_{v^m}^m(a)$ & Set of candidate features to select by agent $a$ in omics $m$ \\
$\mathcal{M}$ & Set of omics \\
$\mathcal{N}_r(u)$ & Neighbors of node $u$ under relation $r$\\
$\rho_\mathcal{V},\rho_\mathcal{E},\rho_\mathcal{M}\in(0,1]$ & Decay coefficients for the dynamic node score, dynamic edge weight, and omics importance \\
$\mathcal{S}_\mathcal{V}, \mathcal{S}_\mathcal{E}$ & Selected features and edges in the current iteration \\
$\mathcal{S}(a)$ & Solution constructed by agent $a$ \\
$\mathcal{S}_\mathcal{V}(a), \mathcal{S}_\mathcal{E}(a)$ & Selected features and edges by agent $a$ in $\mathcal{S}(a)$ \\
$\tau_v^m(t), \tau_{vu}^m(t)$ & Dynamic node score of $f_v$ and dynamic edge weight between $f_v$ and $f_u$ at time $t$ \\
$\boldsymbol{T}_u$ & Cross-modality discovery tensor for patient $u$\\
$\mathcal{W}=\{w_1, w_2,\cdots, w_{M}\}$ & Set of omics importance values \\
${\hat{\mathbf{y}}}_u^m$ & GAT class probability prediction in omics $m$\\
\bottomrule 
\end{tabular}}
\end{table}

\newpage

\section{Implementation Details} 
\label{app:imp} 
HeteroGATomics has been developed using Python 3.10 and PyTorch Geometric 2.4.0 \cite{FeyLenssen2019}, incorporating essential functionalities from PyTorch 2.1.0 \cite{paszke2019pytorch}, scikit-learn 1.3.0 \cite{scikitlearn2011}, NumPy 1.26.0 \cite{harris2020array}, pandas 2.1.1 \cite{reback2023pandas}, and SciPy 1.11.3 \cite{2020SciPy-NMeth}. The training process for the GAT module and VCDN utilizes the PyTorch Lightning 2.1.3 \cite{Falcon_PyTorch_Lightning_2023}. For the training of omics-specific encoders using the GAT module and the VCDN, the Adam optimizer \cite{kingma2015adam} is employed with the StepLR learning rate scheduler strategy, which reduces the learning rate by a factor of 0.8 every 20 epochs. Each omics-specific encoder comprises a three-layer GAT model with hidden dimensions set to [100, 100, 50], incorporating LeakyReLU nonlinearity with a negative slope of 0.01 after each layer. The decoders for each omics type utilize a single-layer fully connected neural network to map the final 50-neuron hidden layer to the output labels. The omics-specific encoder-decoder pairs is pre-trained for 500 epochs, followed by a full training of the entire architecture for an additional 500 epochs. 

When performing baseline feature selection methods, we remove five features at each iteration in RFE, employ the recommended hyperparameter values for mRMR-mv as specified in its original paper, and use the values listed in Table \ref{tbl:common_hyper} for MAgentOmics, ensuring a uniform comparison with HeteroGATomics\textsubscript{MAS}. We utilize the scikit-feature package \cite{li2018feature} for implementing mRMR, and scikit-learn for implementing MI and RFE. 

For classification tasks, we use scikit-learn for KNN, MLP, RF, and Ridge with their default settings, except for the number of iterations in MLP, which is set to 500 epochs. XGBoost is implemented using the XGBoost package \cite{chen2016xgboost}, with its default configurations. For MOGONET, MOSGAT, and {\color{blue}PEARL}, we follow the hyperparameter recommendations from their original publications. To ensure a fair comparison, {\color{red}all three methods} are run for 500 epochs of pre-training for each omics-specific model, followed by an additional 500 epochs for training the entire architecture, similar to the HeteroGATomics setup.

\section{Hyperparameter Setting}
\label{app:hyperparameter}
\begin{figure}[ht!]
\color{blue}
\centering
    \includegraphics[width=1.0\textwidth]{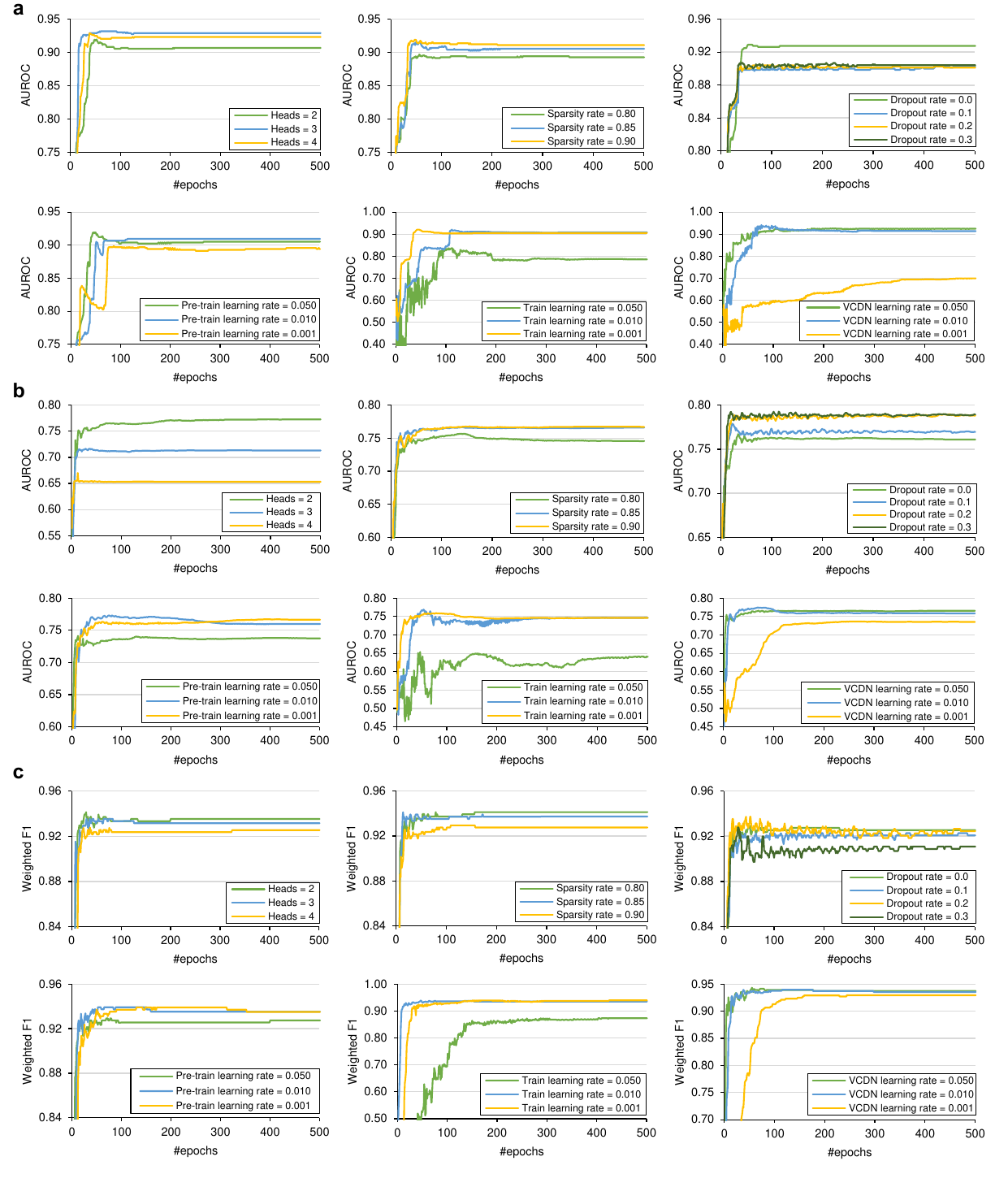}
    \caption{HeteroGATomics classification performance across datasets with different hyperparameter configurations. (a) Results for the BLCA dataset with AUROC. (b) Results for the LGG dataset with AUROC. (c) Results for the RCC dataset with Weighted F1. Averaged values from 10-fold cross-validation are presented. Each fold divides the training set into training and validation sets, with the respective evaluation metric reported for the validation set.}
    \label{fig:hyper_tuning}
\end{figure}

We evaluate six hyperparameters of the GAT module within HeteroGATomics across various datasets, {\color{blue}with results provided in Fig.~\ref{fig:hyper_tuning}}. This analysis uses the average of evaluation metrics from a 10-fold cross-validation across the entire dataset, where, in each fold, the training set is split into training and validation sets at a 9:1 ratio. The results presented are the average values of the evaluation metrics derived from the validation sets. For binary classification tasks in the BLCA and LGG datasets, we utilize AUROC, while Weighted F1 is used for multi-class classification in the RCC dataset. The six evaluated hyperparameters include the number of attention heads, sparsity rate, dropout rate, pre-training learning rate (pre-training lr), training learning rate (training lr), and VCDN learning rate (VCDN lr). Table \ref{tbl:spec_hyper} summarizes the range of possible values for each hyperparameter and highlights the optimal values identified through the optimization process. To assess the impact of individual hyperparameters, each one is varied at a time while keeping the others constant.

\begin{table}[ht!]
\centering
\small
\caption{Dataset-specific hyperparameter configurations for HeteroGATomics.}
\label{tbl:spec_hyper}
\setlength{\tabcolsep}{16pt}
\begin{tabular}{llccc}
\toprule
Hyperparameter     & Range$^{*}$      & BLCA & LGG & RCC \\ 
\midrule 
Number of heads  & \{\textbf{2}, 3, 4\}             & 3             & 2            & 2            \\
Sparsity rate    & \{\textbf{0.80}, 0.85, 0.90\}    & 0.90          & 0.85         & 0.80         \\
Dropout rate     & \{\textbf{0.0}, 0.1, 0.2, 0.3\}  & 0.0           & 0.3          & 0.0          \\
Pre-training lr  & \{0.050, \textbf{0.010}, 0.001\} & 0.010         & 0.001        & 0.001        \\
Training lr      & \{0.050, 0.010, \textbf{0.001}\} & 0.001         & 0.001        & 0.001        \\
VCDN lr          & \{0.050, \textbf{0.010}, 0.001\} & 0.05          & 0.05         & 0.05         \\
\bottomrule 
\end{tabular}
\begin{minipage}{\textwidth}
\vspace{0.1cm}
\footnotesize
$^{*}$Values in sets present the potential options for each hyperparameter, with the bolded value within each set marking the predetermined choice for that hyperparameter when varying another.
\end{minipage}
\end{table}

For the HeteroGATomics\textsubscript{MAS} module, we primarily adopt the hyperparameter values recommended by the foundational MAS algorithm papers \cite{dorigo1997ant, dorigo2005ant, dorigo1999ant}, established through extensive studies. Furthermore, the sparsity rate for each modality within the feature selection module is carefully chosen to guarantee enough edges in the graph, enabling effective traversal by the agents. The common hyperparameter settings applied across all datasets are outlined in Table \ref{tbl:common_hyper}.

\begin{table}[ht!]
\centering
\footnotesize
\caption{Common hyperparameter configurations for HeteroGATomics across datasets.}
\label{tbl:common_hyper}
\setlength{\tabcolsep}{3.5pt}
\begin{tabular}{llr}
\toprule
Module & Hyperparameter & \multicolumn{1}{c}{Value} \\ 
\midrule
HeteroGATomics\textsubscript{MAS} & Number of iterations & 50 \\  
 & Number of agents per omics & 10 \\ 
 & Initial dynamic node score \((c_\mathcal{V})\) & 0.2 \\ 
 & Initial dynamic edge weight \((c_\mathcal{E})\) & 0.2 \\  
 & Node decay coefficient \((\rho_\mathcal{V})\) & 0.1 \\
 & Edge decay coefficient \((\rho_\mathcal{E})\) & 0.1 \\  
 & Omics importance decay coefficient \((\rho_\mathcal{M})\) & 0.1 \\ 
 & Number of selected features in each iteration & 30 \\ 
 & Control parameter \(q_0\) & 0.8 \\ 
 & Sparsity rate in feature similarity networks (DNA, mRNA, miRNA) & [0.9, 0.9, 0.8] \\ 
 \midrule
HeteroGATomics\textsubscript{GAT} & Number of epochs for pre-training model & 500 \\  
 & Number of epochs for training model & 500 \\
 & Number of layers & 3 \\ 
 & Hidden node dimensions for each layer & [100, 100, 50] \\ 
 \bottomrule
\end{tabular}
\end{table}

{\color{blue}
\section{Computational Cost and Scalability Analysis}
\label{app:time-analysis} 
\subsection{Computational Complexity}
\label{app:com_complexity}
Let $N$ denote the number of patients, $M$ the number of omics modalities, and $d_m$ the number of preprocessed features in modality $m$. HeteroGATomics consists of two stages with different computational costs: stage I performs multimodal feature selection using the HeteroGATomics\textsubscript{MAS} module, and stage II performs heterogeneous graph learning using the HeteroGATomics\textsubscript{GAT} module.

\textbf{HeteroGATomics\textsubscript{MAS} module.} The initial construction of the feature similarity networks requires computing relationships between feature pairs within each modality and therefore scales approximately as $\mathcal{O}(\sum_{m=1}^{M} d_m^2)$, before sparsification is applied to reduce the number of retained edges. The subsequent multi-agent search is performed for $T$ iterations using $N_A$ agents per modality. If $C_{\mathrm{search}}$ denotes the cost of constructing and evaluating one candidate feature subset, this step scales as $\mathcal{O}(T \cdot M \cdot N_A \cdot C_{\mathrm{search}})$. Candidate solutions generated by different agents can be evaluated independently, allowing parallel execution. Importantly, the final number of selected features $B$ has limited influence on computational cost because the top $B$ features are selected only after completion of the iterative search. Overall, the computational cost of HeteroGATomics\textsubscript{MAS} is $\mathcal{O}\big(\sum_{m=1}^{M} d_m^2 + T \cdot M \cdot N_A \cdot C_{\mathrm{search}}\big)$, and this module is run once for each training set as a preprocessing step before model training.

\textbf{HeteroGATomics\textsubscript{GAT} module.} After feature selection, each modality $m$ retains $B_m < d_m$ features, with $B=\sum_{m=1}^{M} B_m$. Constructing the patient similarity networks requires pairwise comparisons between patients within each modality and scales as $\mathcal{O}(M \cdot N^2)$, which remains feasible due to the limited number of patients. The feature similarity networks reuse the edge weights already computed in HeteroGATomics\textsubscript{MAS} and therefore add negligible cost. Both networks are constructed once before training. During training, each modality-specific GAT encoder is trained for $T_{\mathrm{pre}} + T_{\mathrm{train}}$ epochs. If $C_{\mathrm{GAT}}$ denotes the cost of one training epoch on a heterogeneous graph, this step scales as $\mathcal{O}\big((T_{\mathrm{pre}} + T_{\mathrm{train}}) \cdot M \cdot C_{\mathrm{GAT}}\big)$. Since sparsification limits the number of patient--patient and feature--feature edges, $C_{\mathrm{GAT}}$ is typically dominated by the $N \cdot B_m$ feature--patient edges and therefore increases with both the number of patients and the number of selected features. Overall, the computational cost of HeteroGATomics\textsubscript{GAT} is $\mathcal{O}\big(M \cdot N^2 + (T_{\mathrm{pre}} + T_{\mathrm{train}}) \cdot M \cdot C_{\mathrm{GAT}}\big)$.

In practice, HeteroGATomics\textsubscript{MAS} accounts for most of the computational cost, while HeteroGATomics\textsubscript{GAT} remains efficient owing to the reduced feature space (\ref{app:scalability}).
}

\subsection{Runtime and Scalability}
\label{app:scalability}
\begin{figure}[h]
\centering
    \includegraphics[width=0.9\textwidth]{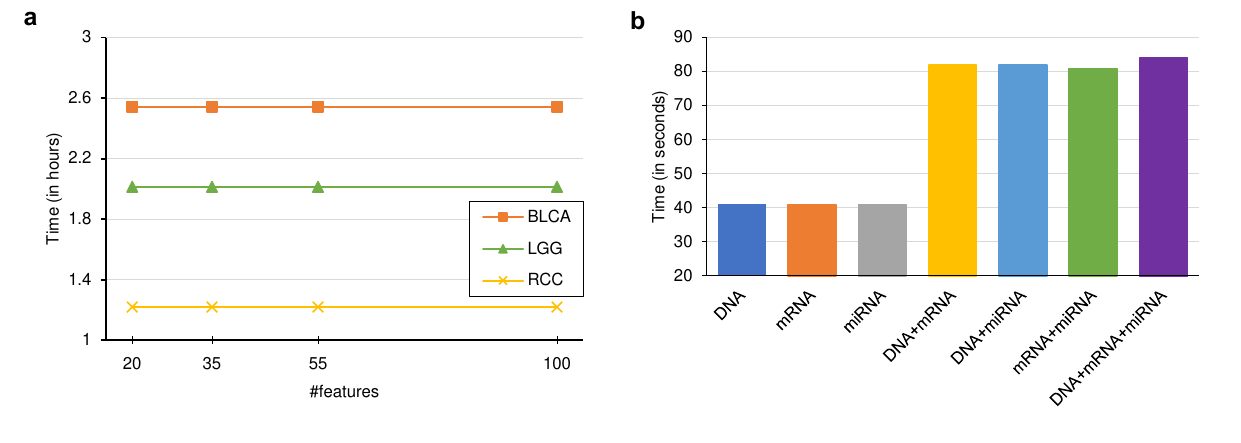}
    \caption{Average execution times per fold over 10-fold cross-validation. (a) Execution time of HeteroGATomics\textsubscript{MAS} for different numbers of selected features across three datasets. (b) Execution time of HeteroGATomics\textsubscript{GAT} for different combinations of omics modalities on LGG.}
    \label{fig:time_analysis}
\end{figure}

We measure the execution time of the two modules of HeteroGATomics: HeteroGATomics\textsubscript{MAS} and HeteroGATomics\textsubscript{GAT}. For HeteroGATomics\textsubscript{MAS}, we measure only the time required for feature selection, independent of any downstream classifier. For HeteroGATomics\textsubscript{GAT}, the execution time includes the construction of heterogeneous graphs, pre-training of each omics-specific GAT model, training of each omics-specific GAT with the VCDN module (inactive for a single modality), and testing of the entire architecture. HeteroGATomics\textsubscript{MAS} is run on a 20-core CPU on the Stanage High-Performance Computing cluster at The University of Sheffield, and HeteroGATomics\textsubscript{GAT} on an NVIDIA GeForce RTX 4090 GPU. The results are presented in Fig.~\ref{fig:time_analysis}.

Figure~\ref{fig:time_analysis}(a) shows that the execution time of HeteroGATomics\textsubscript{MAS} ranges from approximately 1.2 to 2.5 hours across datasets and remains nearly constant as the number of selected features varies. This is consistent with \ref{app:com_complexity}, as the top $B$ features are selected only after the iterative search is completed. In contrast, execution time increases with the total number of features in the dataset, with BLCA requiring the longest time and RCC the shortest.

Figure~\ref{fig:time_analysis}(b) shows that HeteroGATomics\textsubscript{GAT} requires 84 seconds when all three omics modalities are used. The execution time differs only slightly between two and three modalities, indicating that adding a modality incurs a small additional cost. The single-modality configurations require less time because the VCDN module is inactive during training.

{\color{blue}
\subsection{Computational Time Comparison}
To compare the computational cost of HeteroGATomics with existing methods, Fig.~\ref{fig:time_comp} presents classification performance against average execution time over 10 cross-validation folds for BLCA, LGG, and RCC. Execution time is shown on a base-10 logarithmic scale because the evaluated methods span several orders of magnitude in runtime. The two modules of HeteroGATomics are reported separately because they have substantially different computational characteristics: HeteroGATomics\textsubscript{MAS} performs the iterative multi-agent feature selection, whereas HeteroGATomics\textsubscript{GAT} performs the subsequent heterogeneous graph learning and multimodal integration.

\begin{figure}[h]
\color{blue}
\centering
    \includegraphics[width=1.0\textwidth]{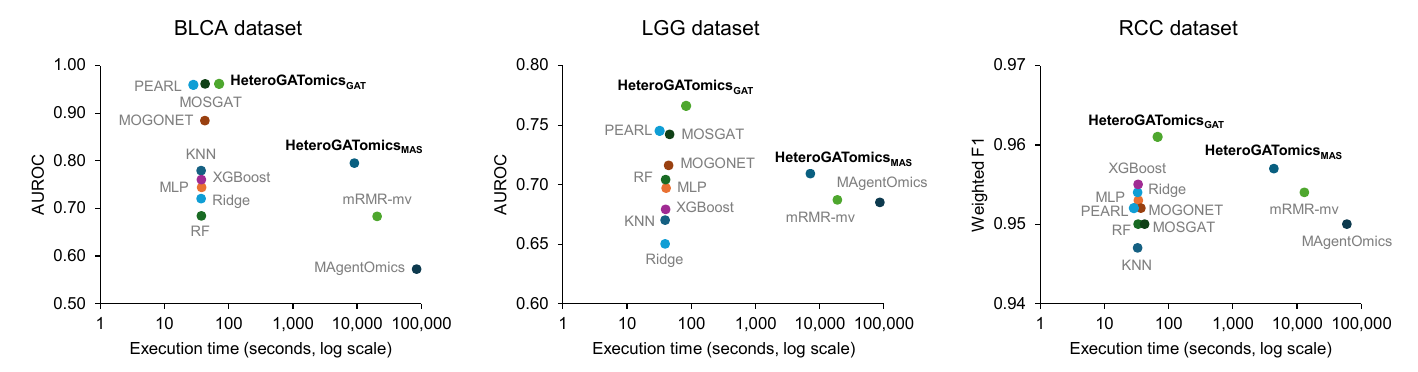}
    \caption{Comparison of classification performance and execution time across methods, averaged over 10 cross-validation folds. Execution time is shown on a base-10 logarithmic scale.}
    \label{fig:time_comp}
\end{figure}
As shown in Fig.~\ref{fig:time_comp}, conventional classifiers combined with MI require the lowest execution times, generally around 30--40 seconds. Graph-based methods also remain computationally efficient: PEARL requires approximately 29--33 seconds owing to its simple spectral graph convolution, while MOGONET and MOSGAT require approximately 37--47 seconds. HeteroGATomics\textsubscript{GAT} requires approximately 63--84 seconds while achieving the highest predictive performance among all methods on LGG and RCC and the joint-highest on BLCA, indicating a favourable balance between computational cost and classification performance. In contrast, feature selection methods involving iterative search require substantially longer execution times, reflecting the cost of exploring the high-dimensional multimodal feature space. Notably, HeteroGATomics\textsubscript{MAS} is considerably faster than both multimodal feature selection baselines, mRMR-mv and MAgentOmics, while achieving higher predictive performance. Moreover, as discussed in Section~\ref{app:time-analysis}, the multi-agent search can be executed in parallel, further reducing the practical wall-clock time of this module.
}

{\color{blue}
\subsection{Memory and Parameter Analysis}
\label{app:memory}
Table~\ref{tbl:gat_resources} reports the number of trainable parameters and peak GPU memory of HeteroGATomics\textsubscript{GAT} across the three datasets. HeteroGATomics\textsubscript{GAT} contains approximately 1.13--1.50 million trainable parameters and requires approximately 0.9--1.2 GB of peak GPU memory, indicating that it can be trained on standard GPU hardware. In contrast, HeteroGATomics\textsubscript{MAS} has no trainable parameters, as it is a search-based feature selection method.

\begin{table}[h]
\color{blue}
\centering
\setlength{\tabcolsep}{1.5mm}
\begin{tabular}{lll}
\hline
Dataset & \#Parameters & Peak GPU memory (MB) \\
\hline
BLCA & 1.50 M & 868 \\
LGG  & 1.13 M & 952 \\
RCC  & 1.17 M & 1233 \\
\hline
\end{tabular}
\caption{Number of trainable parameters and peak GPU memory of HeteroGATomics\textsubscript{GAT}, averaged over 10-fold cross-validation.}
\label{tbl:gat_resources}
\end{table}
}

{\color{blue}
\section{Statistical Significance Analysis}
\label{app:significance}
To assess whether the performance differences between HeteroGATomics and the baseline methods reported in Tables~2--4 are statistically meaningful, we perform two-sided Wilcoxon signed-rank tests \cite{demvsar2006statistical} on the per-fold results of the 10-fold cross-validation, comparing HeteroGATomics with each baseline on each evaluation metric. Since all methods are evaluated on identical splits, the tests are paired by fold. The results are shown in Fig.~\ref{fig:significance}.

\begin{figure}[h]
\color{blue}
\centering
    \includegraphics[width=1.0\textwidth]{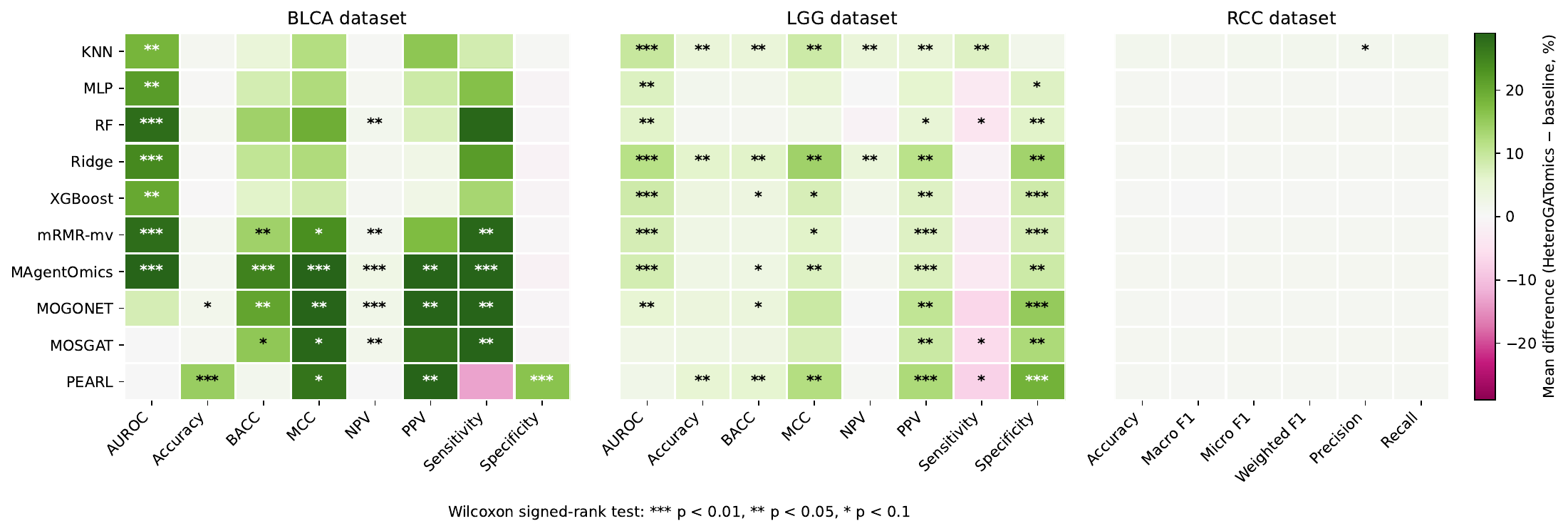}
    \caption{Statistical comparison of HeteroGATomics with each baseline based on the results in Tables~2--4. Color shows the mean difference in performance over the 10 cross-validation folds (green: HeteroGATomics is better; pink: the baseline is better). Stars denote the significance of a two-sided Wilcoxon signed-rank test across folds (***: $p < 0.01$, **: $p < 0.05$, *: $p < 0.1$).}
    \label{fig:significance}
\end{figure}

On BLCA and LGG, statistically significant improvements of HeteroGATomics are observed over all conventional classifiers and multimodal feature selection methods in one or more evaluation metrics. Statistically significant improvements are also observed over the graph-based methods MOGONET, MOSGAT, and PEARL in several metrics, such as MCC, PPV, and specificity.

The pattern of non-significant or unfavourable comparisons is consistent with the characteristics of the datasets and metrics. On LGG, the lower sensitivity of HeteroGATomics is accompanied by significantly higher specificity against most baselines, reflecting a different balance between the two classes rather than weaker overall classification. This is supported by balanced metrics that account for both classes: compared with PEARL, which achieves higher sensitivity, HeteroGATomics achieves significantly higher BACC, MCC, accuracy, PPV, and specificity. On BLCA, specificity is close to the ceiling for most methods owing to the large majority of high-grade patients, resulting in generally small differences, except for PEARL, which HeteroGATomics significantly outperforms in specificity. On RCC, all methods achieve performance close to the ceiling, as the RCC subtypes are relatively easy to distinguish \cite{wang2021mogonet}, leaving little room for significant differences, although HeteroGATomics achieves the highest mean performance across all six metrics.
}

\section{Biomarker Identification and Interpretation Techniques}
\label{app:biomarker-tech} 
HeteroGATomics facilitates the identification of key cancer biomarkers, enhancing our understanding of its decision-making process. This process is essential for interpreting the architecture's effectiveness and pinpointing the most informative features serving as cancer biomarkers. We leverage an ablation approach commonly used in deep learning-based methods for biomarker selection \cite{wang2021mogonet,setiono1997neural,amjad2022neural}. This approach involves systematically removing each feature to observe its impact on the model's performance. Specifically, for each feature within a given omics, we temporarily remove the feature by setting its value and its attributes (i.e., static and dynamic node scores) to zero in both the feature and patient similarity networks. Then, we evaluate the model's classification performance on the test set without this feature. Features whose removal leads to the most significant decrease in classification performance are considered top biomarkers. This decrease indicates the feature's substantial impact on the model, highlighting its importance. We apply a 10-fold cross-validation strategy, indicating that the set of input features to the GAT module may vary across different folds. Therefore, we assess the model's classification performance for each feature in every fold. To rank features from each omics modality, we measure the cumulative reduction in classification performance, which is then normalized according to the frequency of their occurrences across all folds. To evaluate the performance of HeteroGATomics, we use AUROC for binary classification tasks in the BLCA and LGG datasets.

GO enrichment analyses are performed with GOseq in R, using GO biological process (GO BP),  molecular function (MF), and cellular component (CC). The top 30 biomarkers for mRNA and DNA omics categories are used as enrichment sets with the remaining 300 biomarkers as the background set. Protein-protein interaction networks are generated in Cytoscape, using string-db \cite{szklarczyk2023string} and ConsensusPathway \cite{kamburov2022consensuspathdb} databases for known physical interactions. Only direct two-by-two interactions are used, with an interaction confidence of $>=$ 7 for string-db and $>=$ 9.5 for ConsensusPathway (reported as high confidence interactions). Known cancer-related genes are fetched from the Cancer Gene Census database \cite{sondka2024cosmic}, OncoKB™ Cancer Gene List \cite{chakravarty2017oncokb}, and the Network of Cancer Genes \cite{repana2019ncg}. Target mRNAs for miRNA biomarkers are inferred from starBase \cite{li2014starbase}, keeping only targets with at least one CLIP experiment evidence and predicted by at least two miRNA target predictor tools (e.g. miRanda and/or TargetScan).

\section{Additional Analysis of Identified Biomarkers}
\begin{figure}[!t]
\centering
    \includegraphics[width=0.95\textwidth, trim=4cm 0 4cm 0]{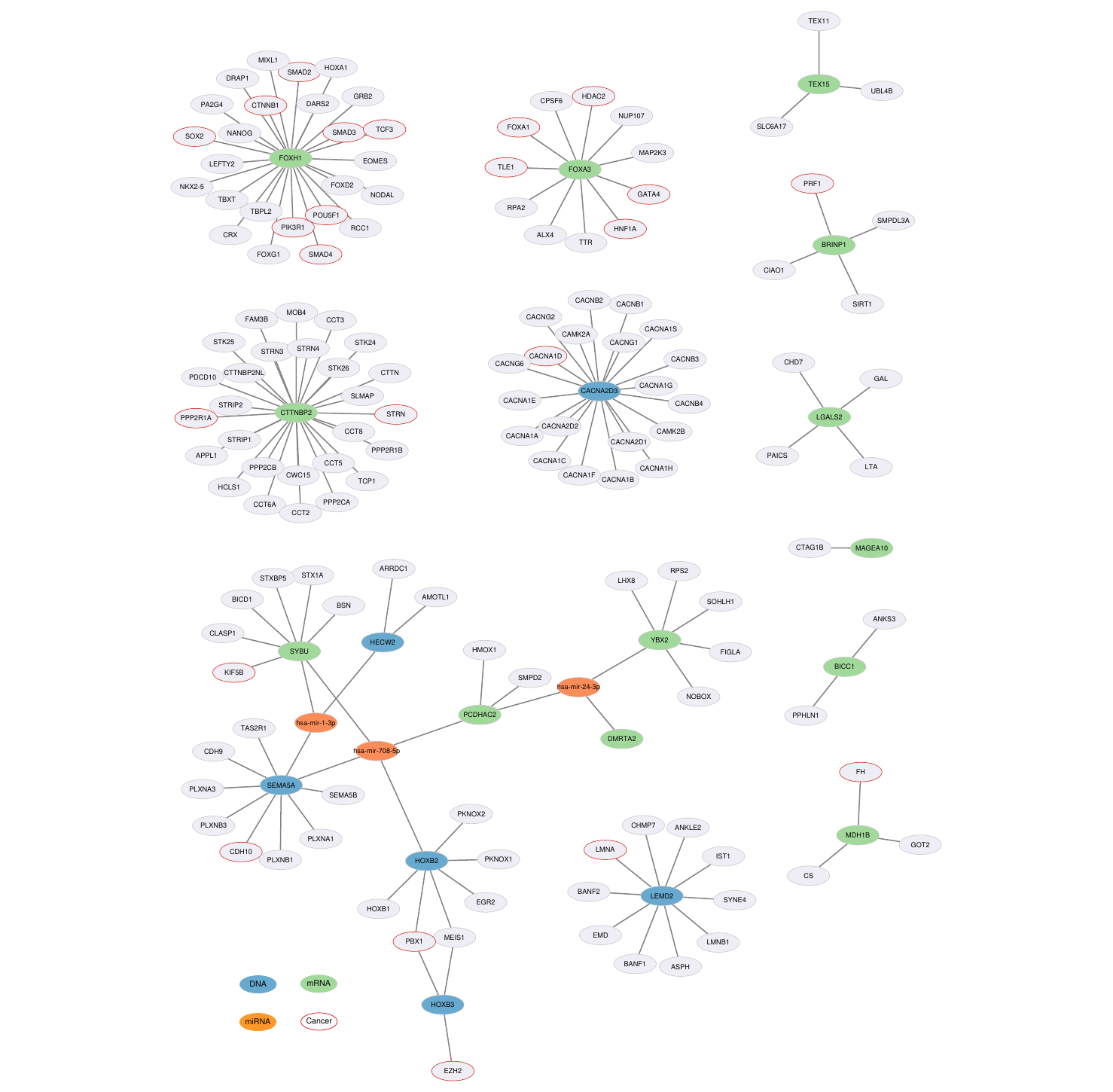}
    \caption{Interaction network of top 30 biomarkers with known partners for BLCA. Direct protein-protein interactions are recovered for DNA and mRNA omics. For the miRNA omics, known mRNA targets are recovered from starBase \cite{li2014starbase}.}
    \label{fig:partners_blca}
\end{figure}

\begin{figure*}[!t]
\centering
    \includegraphics[width=0.98\textwidth, trim=4cm 0 4cm 1cm]{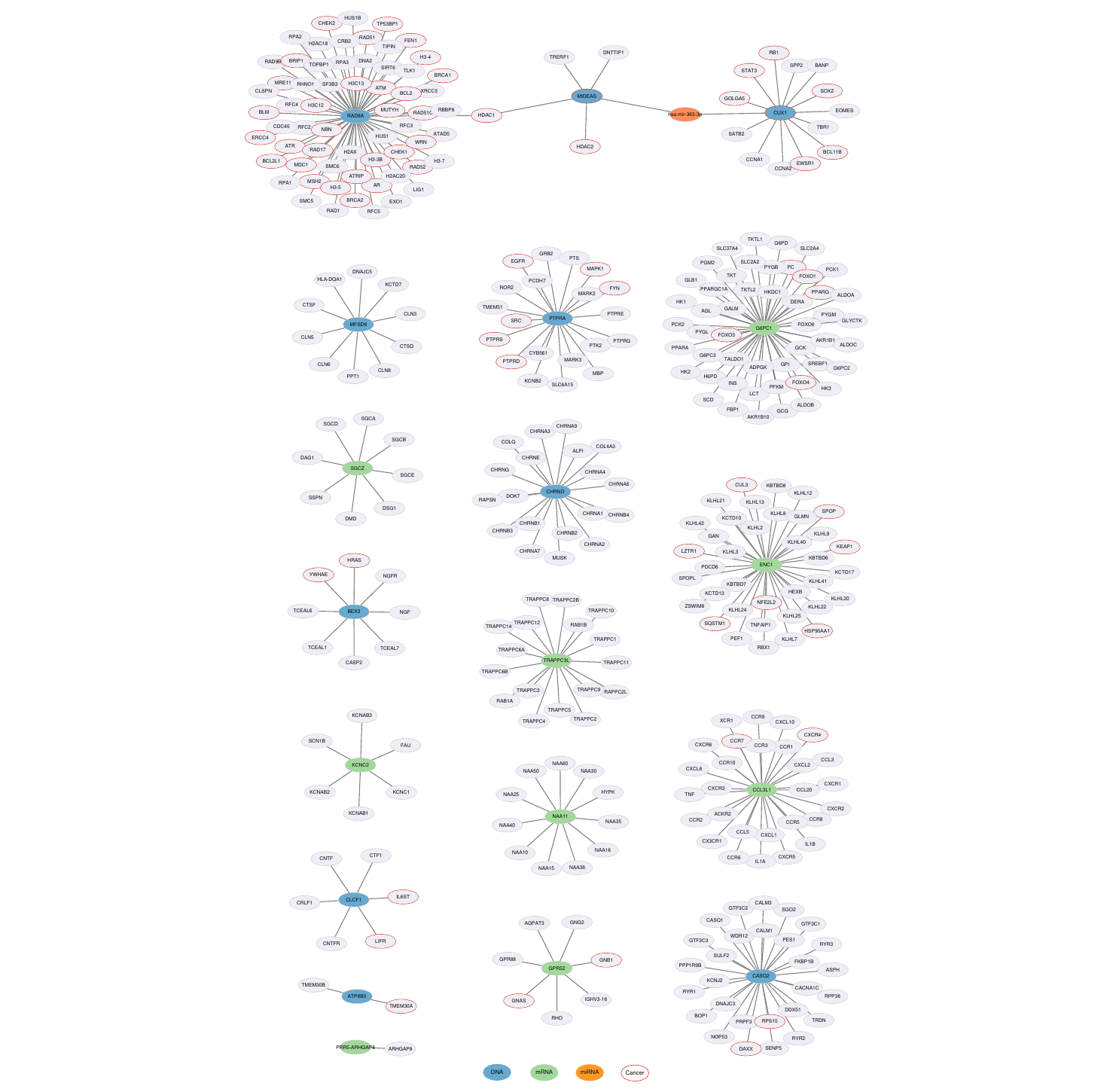}
    \caption{Interaction network of top 30 biomarkers with known partners for LGG. Direct protein-protein interactions are recovered for DNA and mRNA omics. For the miRNA omics, known mRNA targets are recovered from starBase \cite{li2014starbase}.}
    \label{fig:partners_lgg}
\end{figure*}

\begin{figure}[!t]
\centering
    \includegraphics[width=0.9\textwidth]{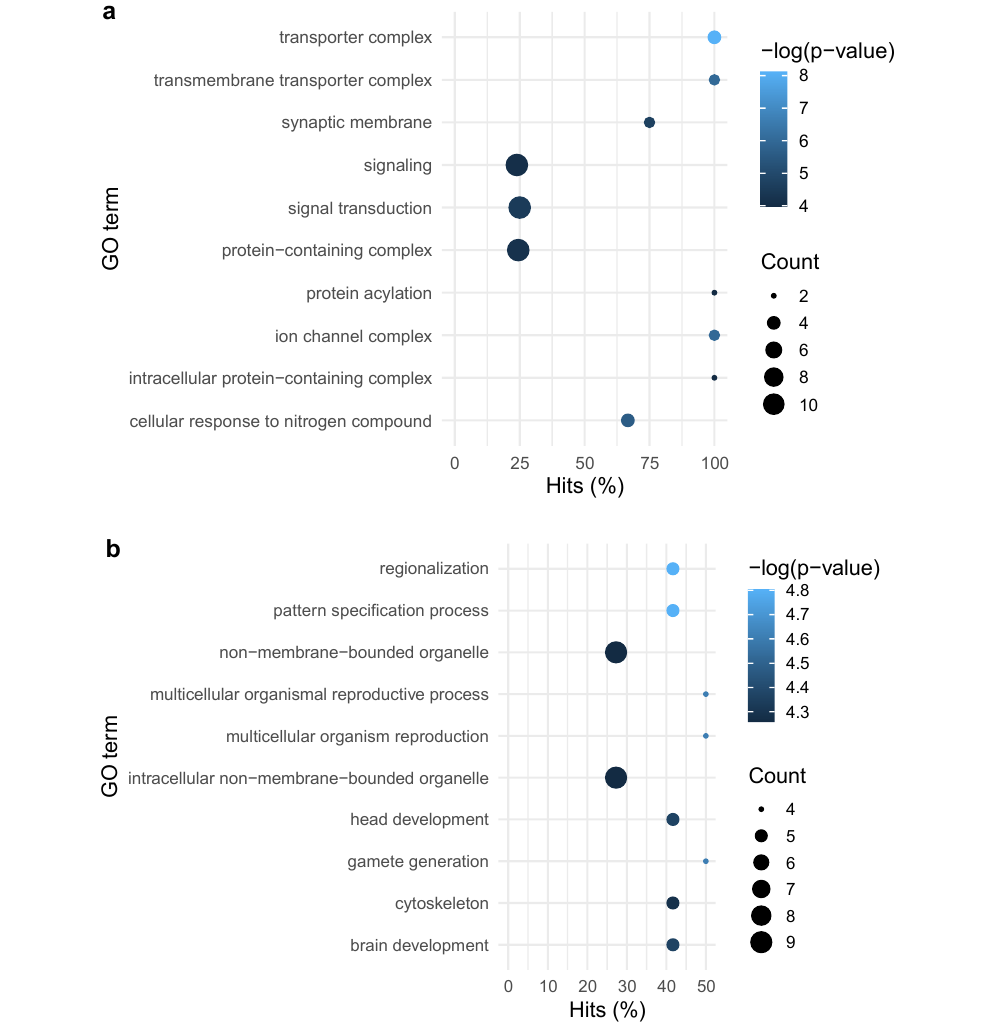}
    \caption{GO enrichment analysis of the top 30 biomarkers from the DNA and mRNA omics modalities for (a) LGG and (b) BLCA. The y-axis shows the top 10 most significant GO category terms, while the x-axis represents the percentage of biomarkers belonging to each GO category.}
    \label{fig:go}
\end{figure}

Tables \ref{tbl:biomarker_blca} and \ref{tbl:biomarker_lgg} present the 30 most important biomarkers identified by HeteroGATomics on BLCA and LGG, respectively. We provide further biological interpretation of the biomarkers identified by HeteroGATomics, as shown in Figs. \ref{fig:partners_blca}, \ref{fig:partners_lgg}, and \ref{fig:go}.

{\color{blue}Since biomarker importance is evaluated on the features selected by HeteroGATomics\textsubscript{MAS} in each fold, we assess the stability of these candidate biomarkers by counting the number of folds in which each feature is selected. Of the 300 features selected per fold, 175 (58\%) on LGG and 179 (60\%) on BLCA are consistently selected in all 10 folds, and 217 (72\%) and 211 (70\%), respectively, are selected in at least 8 folds. These results indicate that HeteroGATomics\textsubscript{MAS} consistently identifies a core set of candidate biomarkers across different training sets.}

\begin{table}[h]
\centering
\small
\caption{Top 30 ranked biomarkers identified by HeteroGATomics in the BLCA dataset.}
\label{tbl:biomarker_blca}
\setlength{\tabcolsep}{8mm}
{\begin{tabular}{llll}
\toprule
Rank & Biomarker ID & Biomarker name$^{*}$  & Omics \\
\midrule
1 & MIMAT0000416 & hsa-mir-1-3p & miRNA \\
2 & MIMAT0004926 & hsa-mir-708-5p & miRNA \\
3 & MIMAT0004518 & hsa-mir-16-2-3p & miRNA \\
4 & MIMAT0009451 & hsa-mir-1976 & miRNA \\
5 & MIMAT0000080 & hsa-mir-24-3p & miRNA \\
6 & DBC1 & BRINP1 & mRNA \\
7 & cg12676289 & CTD-2201E9.2,SEMA5A & DNA \\
8 & cg22777724 & HOXB2,HOXB-AS1	& DNA \\
9 & cg26681383 & CACNA2D3 & DNA \\
10 & cg09313705 & HOXB2,HOXB-AS1 & DNA \\
11 & cg20152430 & HOXB-AS3,HOXB3 & DNA \\
12 & LGALS2 & LGALS2 & mRNA \\
13 & MAGEA10 & MAGEA10 & mRNA \\
14 & MDH1B & MDH1B & mRNA \\
15 & FOXH1 & FOXH1 & mRNA \\
16 & YBX2 & YBX2 & mRNA \\
17 & PCDHAC2 & PCDHAC2 & mRNA \\
18 & LOC644172 & LOC644172 & mRNA \\
19 & CTTNBP2 & CTTNBP2 & mRNA  \\
20 & DMRTA2 & DMRTA2 & mRNA  \\
21 & TEX15 & TEX15 & mRNA \\
22 & SYBU & SYBU & mRNA \\
23 & BICC1 & BICC1 & mRNA \\
24 & FOXA3 & FOXA3 & mRNA \\
25 & cg27452922 & N/A & DNA	\\
26 & cg11241756 & H2BP2 & DNA \\
27 & cg20197814 & HECW2 & DNA \\
28 & cg00334056 & LEMD2	& DNA \\
29 & cg22968622 & DND1P1 & DNA \\
30 & cg08836615 & N/A & DNA \\
\bottomrule
\end{tabular}}
\begin{minipage}{\textwidth}
\vspace{0.1cm}
\footnotesize
$^{*}$ N/A indicates the biomarker name for the corresponding ID is not available.
\end{minipage}
\end{table}

\begin{table}[t]
\centering
\small
\caption{Top 30 ranked biomarkers identified by HeteroGATomics in the LGG dataset.}
\label{tbl:biomarker_lgg}
\setlength{\tabcolsep}{4mm}
{\begin{tabular}{llll}
\toprule
Rank & Biomarker ID & Biomarker name$^{*}$  & Omics \\
\midrule
1	& cg15024277 & BEX3 & DNA  \\
2 & cg20371266 & CHRND & DNA  \\
3 & cg22373770 & N/A & DNA  \\
4	& cg05165025 & MIDEAS,RP5-1021I20.1 & DNA  \\
5 & cg16503259 & N/A	& DNA  \\
6 & cg20253855 & CUX1 & DNA  \\
7	& cg17237063 & RBMS3-AS3,RBMS3 & DNA  \\
8 & cg15275625 & N/A & DNA  \\
9	& RNF126P1 & RNF126P1 & mRNA \\
10 & TTTY14 & TTTY14 & mRNA \\
11 & C4orf45 & SPMIP2 & mRNA \\
12 & SGCZ & SGCZ & mRNA \\
13	& NAA11 & NAA11	& mRNA \\
14 & cg00661753 & PTPRA & DNA  \\
15 & BET3L & TRAPPC3L & mRNA \\
16 & ZDHHC1 & ZDHHC1 & mRNA \\
17 & G6PC & G6PC1 & mRNA \\
18 & GPR52 & GPR52 & mRNA \\
19 & cg12472597 & CTC-1337H24.4,CLCF1,RAD9A,AP003419.11	& DNA  \\
20 & cg11867599 & ABHD18,MFSD8 & DNA  \\
21 & CCL3L1 & CCL3L1 & mRNA \\
22 & cg00020474 & N/A & DNA  \\
23 & cg14302471 & LINC00689 & DNA  \\
24 & ENC1 & ENC1 & mRNA \\
25 & PRR5-ARHGAP8	& PRR5-ARHGAP8 & mRNA \\
26 & cg14985891 & CASQ2 & DNA  \\
27 & cg07971493 & MAP3K15 & DNA  \\
28 & cg08316083 & ATP8B3 & DNA  \\
29	& MIMAT0000707 & hsa-mir-363-3p	& miRNA \\
30 & KCNC2 & KCNC2 & mRNA \\
\bottomrule
\end{tabular}}
\begin{minipage}{\textwidth}
\vspace{0.1cm}
\footnotesize
$^{*}$ N/A indicates the biomarker name for the corresponding ID is not available.
\end{minipage}
\end{table}

\clearpage

\putbib[main]
\end{bibunit}
\label{ref:sup}

\end{document}